# TemporalPaD: a reinforcement-learning framework for temporal feature representation and dimension reduction

Xuechen Mu[1,2], Zhenyu Huang[1,3], Kewei Li[1,3], Haotian Zhang[1,3], Xiuli Wang[2], Yusi Fan[1,3], Kai Zhang[1,2,*], Fengfeng Zhou[1,3,4,*].

1 Key Laboratory of Symbolic Computation and Knowledge Engineering of Ministry of Education, Jilin University, Changchun 130012, Jilin, China.

2 School of Mathematics, Jilin University, Changchun 130012, Jilin, China.

3 College of computer Science and Technology, Jilin University, Changchun, Jilin, China, 130012.

4 School of Biology and Engineering, Guizhou Medical University, Guiyang 550025, Guizhou, China

* Correspondence may be addressed to Fengfeng Zhou: FengfengZhou@gmail.com or ffzhou@jlu.edu.cn. Correspondence may also be addressed to Kai Zhang: zhangkaimath@jlu.edu.cn.

# Abstract

Recent advancements in feature representation and dimension reduction have highlighted their crucial role in enhancing the efficacy of predictive modeling. This work introduces TemporalPaD, a novel end-to-end deep learning framework designed for temporal pattern datasets. TemporalPaD integrates reinforcement learning (RL) with neural networks to achieve concurrent feature representation and feature reduction. The framework consists of three cooperative modules: a Policy Module, a Representation Module, and a Classification Module, structured based on the Actor-Critic (AC) framework. The Policy Module, responsible for dimensionality reduction through RL, functions as the actor, while the Representation Module for feature extraction and the Classification Module collectively serve as the critic. We comprehensively evaluate TemporalPaD using 29 UCI datasets, a well-known benchmark for validating feature reduction algorithms, through 10 independent tests and 10-fold cross-validation. Additionally, given that TemporalPaD is specifically designed for time series data, we apply it to a real-world DNA classification problem involving enhancer category and enhancer strength. The results demonstrate that TemporalPaD is an efficient and effective framework for achieving feature reduction, applicable to both structured data and sequence datasets. The source code of the proposed TemporalPaD is freely available as supplementary material to this article and at http://www.healthinformaticslab.org/supp/.

# Introduction

The advent of big data technologies and advanced storage solutions has boosted an era with an unprecedented volume of data [1]. This exponential increase in data volume presents not only a challenge for data storage capacity, but also the requirement of effective and accurate computational algorithms to extract valuable information from the vast amount of available data. Feature representation (FR) aims to tackle data redundancy by distilling the most relevant information for predictive models [2, 3], while dimension reduction (DR) is designed to reduce the number of features [4, 5].

FR has found widespread applications across various domains, especially in handling complex datasets. For example, FR plays a critical role in extracting key information for image-based processing tasks [6]. The effective analysis and interpretation of text-based natural language processing tasks can also be enabled by FR [7]. Stock prices and economic trends in the financial sector have been accurately forecasted by efficient FR algorithms [8]. The primary challenge of FR lies in identifying the most relevant insights from large datasets, a task that becomes particularly complex with time-series data [9].

The extraction of relevant information necessitates further reduces the dimensions of the newly-engineered feature space. DR typically involves two main strategies: feature selection and engineering-based dimension reduction. Feature selection tries to recommend a subset of features from the original dataset based on criteria such as relevance and collinearity, thereby reducing the number of feature dimensionality [10-13]. The engineering-based dimension reduction method, on the other hand, involves a more profound transformation of the feature set into a new set of abstract features, which often lack direct interpretations [14]. Popular algorithms include principal component analysis (PCA) [15] and singular value decomposition (SVD) [16]. Neural network-based autoencoders have also been utilized to learn low-dimensional representations of raw data, which fulfills the goal of feature dimension reduction [17-20]. Yet a unified approach for integrating feature representation and dimension reduction in time-series data has been elusive.

To address these challenges, we propose a novel end-to-end deep learning framework called TemporalPaD, designed for temporal pattern datasets. TemporalPaD integrates RL [21] with neural networks to achieve concurrent feature extraction and feature reduction. The framework employs the actor-critic architecture [21, 22], where the Policy Module, responsible for dimensionality reduction through RL, acts as the actor, while the Representation Module for feature extraction and the Classification Module collectively serve as the critic. We comprehensively evaluate TemporalPaD using the UCI dataset, a well-known benchmark for validating feature reduction algorithms, through 10 independent tests and 10-fold cross-validation. Beyond these evaluations, given that TemporalPaD is specifically designed for time series data, we applied it to a real-world DNA classification problem involving enhancer category and enhancer strength. The results demonstrate that TemporalPaD is an efficient and effective framework for achieving feature reduction, applicable to both structured data and sequence datasets.

## Materials and methods

### Datasets

**Table 1.** Detailed descriptions of the datasets employed in this study. The column "Src" indicates that the dataset is a UCI structured dataset or a biological sequence dataset. The numbers of samples, features, and labels of each dataset are given in respective "Samples", "Features", and "Labels" columns. Each sample in the two "Bio" datasets is a DNA sequence with the length of 200 base pairs (bp), and will be transformed into numerical features through the $k$-mer method. The parameter "$k$" will be evaluated in the following section.

| ID | Src | Dataset | Features | Samples | Labels | subject area |
|---|---|---|---|---|---|---|
| 1 | UCI | Arrhythmia | 279 | 452 | 16 | health and medicine |
| 2 | UCI | Austra | 14 | 690 | 2 | business |
| 3 | UCI | Breast | 9 | 699 | 2 | health and medicine |
| 4 | UCI | Chess | 36 | 3196 | 2 | games |
| 5 | UCI | Credit | 15 | 690 | 2 | business |
| 6 | UCI | Diabetes | 8 | 768 | 2 | health and medicine |
| 7 | UCI | German | 20 | 1000 | 2 | social science |
| 8 | UCI | Glass | 9 | 214 | 6 | physics and chemistry |
| 9 | UCI | Heart | 13 | 270 | 2 | health and medicine |
| 10 | UCI | Horse-colic | 26 | 368 | 2 | biology |
| 11 | UCI | Hypothyroid | 25 | 3163 | 2 | health and medicine |

| 12 | UCI | Ionosphere | 34 | 351 | 2 | physics and chemistry |
| 13 | UCI | Iris | 4 | 150 | 3 | biology |
| 14 | UCI | Liver Disorders | 7 | 345 | 2 | health and medicine |
| 15 | UCI | Lung-cancer | 56 | 32 | 3 | health and medicine |
| 16 | UCI | Mushroom | 22 | 8124 | 2 | biology |
| 17 | UCI | Pima | 8 | 768 | 2 | health and medicine |
| 18 | UCI | Segmentation | 19 | 2310 | 7 | other |
| 19 | UCI | Sonar | 60 | 208 | 2 | physics and chemistry |
| 20 | UCI | Spambase | 57 | 4601 | 2 | computer science |
| 21 | UCI | Spect | 22 | 267 | 2 | health and medicine |
| 22 | UCI | Vehicle | 18 | 846 | 4 | other |
| 23 | UCI | Vote | 16 | 435 | 2 | social science |
| 24 | UCI | Waveform | 40 | 5000 | 3 | physics and chemistry |
| 25 | UCI | WDBC | 30 | 569 | 2 | health and medicine |
| 26 | UCI | WPBC | 34 | 198 | 2 | health and medicine |
| 27 | UCI | Wine | 13 | 178 | 3 | physics and chemistry |
| 28 | UCI | Yeast | 8 | 1484 | 10 | biology |
| 29 | UCI | Zoo | 17 | 101 | 2 | biology |
| 30 | Bio | Enhancer1 | 2968 | 200bp | 2 | DNA Sequence |
| 31 | Bio | Enhancer2 | 1484 | 200bp | 2 | DNA Sequence |

Our evaluation of the TemporalPaD framework, along with other comparative algorithms, encompasses two distinct types of datasets (Table 1). The first category comprises 29 widely-used benchmark datasets from the UCI Machine Learning

Repository [23]. These datasets cover a broad spectrum of domains, including biomedicine and natural language processing (NLP). They offer a comprehensive and fair platform for assessing the performance of TemporalPaD and other algorithms across various application areas.

The second category includes two publicly available biological datasets of enhancer classification [24]. The first dataset Enhancer1 consists of a training subset with 2968 samples (1484 enhancers/positives and 1484 non-enhancers/negatives) and an independent test subset with 400 samples (200 enhancers/positives and 200 non-enhancers/negatives). The next dataset Enhancer2 is for enhancer strength classification, consisting of 742 strong (positive) and 742 weak (negative) enhancers in the training subset, and 100 positive and 100 negative samples in the test subset. Each sample is a DNA sequencer with the length of 200 base pairs (bp), and will be transformed into the numerical features through the *k*-mer method [25, 26]

## Performance metrics

The first type of 29 UCI datasets is evaluated using the overall accuracy. This metric is defined as the proportion of correctly classification samples across all class labels in a dataset. A random split was performed to divide the dataset into a training set, which consisted of two-thirds of the total dataset, and a test set, which comprised one-third of the total dataset. Ten random runs are conducted, as similar in [27]. An additional experiment of ten-fold cross validation (10FCV) strategy is also conducted to evaluate the algorithms on these datasets.

The second type consists of two binary classification tasks and is evaluated by the following five performance metrics [28, 29], i.e., accuracy (Acc), sensitivity (Sn), specificity (Sp), Matthews correlation coefficient (MCC), and the area under the ROC curve (AUC). These metrics are defined in the equations (1)-(4).

$$Acc = 1 - \frac{S_-^+ + S_+^-}{S_+ + S_-} \quad (1)$$

$$Sn = 1 - \frac{S_-^+}{S_+} \quad (2)$$

$$Sp = 1 - \frac{S_+^-}{S_-} \quad (3)$$

$$MCC = \frac{1 - \frac{S_-^+ + S_+^-}{S_+ + S_-}}{\sqrt{[1 + \frac{S_+^- - S_-^+}{S_+}][1 + \frac{S_-^+ - S_+^-}{S_-}]}} \qquad (4)$$

Here, $S_+$ represents the number of positive samples, while $S_-$ represents the number of negative samples. Additionally, $S_-^+$ denotes the number of actual negative samples that are incorrectly predicted as positives, while $S_+^-$ refers to the number of actual positive samples that are incorrectly predicted as negatives.

## Overall architecture of TemporalPaD

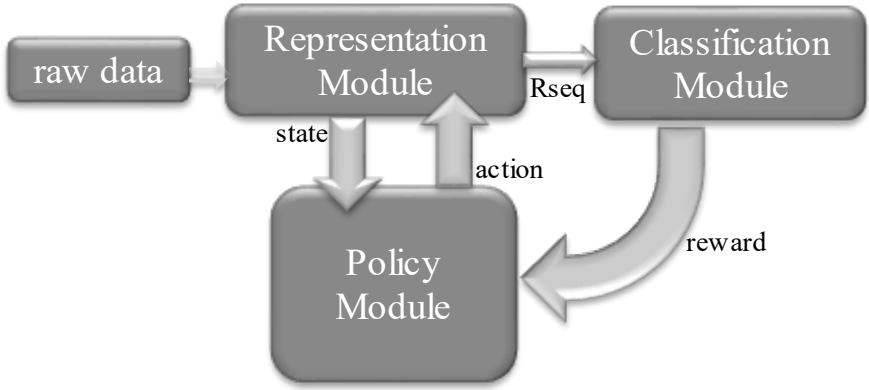

**Figure 1. Flowchart of the proposed TemporalPaD framework.** The framework consists of three main modules: Representation, Policy, and Classification.

This study tackle the Feature Representation (FR) challenge utilizing the actor-critic paradigm within the deep Reinforcement Learning (RL) architecture [21, 30]. The proposed TemporalPaD framework (Figure 1) integrates three core modules, the Policy Module, the Representation Module, and the Classification Module. The Policy Module functions as the "actor" to actively make decisions, whereas the Representation and Classification Modules collectively serve as the "critic" by providing evaluation feedback. This Deep Learning (DL) architecture is employed to formulate the FR challenge as a Markov Decision Process (MDP) with delayed rewards.

Figure 1 illustrates the process wherein data samples are initially transformed into a representation space by the Representation Module. These transformed feature vectors form the elements of the state space, which are subsequently fed into the Policy Module. The Policy Module then makes decisions about whether to retain (denoted as "1") or discard (denoted as "0") the represented features, resulting in a new masked sequence of features (Rseq). The selected features, Rseq, are then passed to the Classification Module. The outcomes of the classification process serve as rewards for the Policy Module, guiding its decision-making process.

The intricate interplay and specific functionalities of these deeply integrated modules are further delineated in subsequent sections.

## Policy module

The policy module of TemporalPaD adopts the standard policy function $\pi(a_t|s_t;\theta)$ commonly used in the RL architecture [31]. This process follows a typical delayed reward paradigm. The policy module utilizes the probability distribution defined by $\pi(a_t|s_t;\theta)$ to select the appropriate action based on the input feature representation vector from the representation module. All the input feature representations are traversed until this process stops.

The newly selected subset of the represented features is then utilized as the input to the classification module. The prediction results of the classification module serve as the delayed rewards to evaluate the decisions made by the policy module. This interaction allows for the continuously monitoring and updating of the policy module. The learned policy function is defined as:

$$\pi(a_t|s_t;\theta) = \sigma(W \times s_t + b) \quad (5)$$

The action taken at time scale $t$ is denoted as $a_t$, while the state value at time scale $t$ is denoted as $s_t$. The network parameters ($W$, $b$) of the policy module are collectively represented by $\theta$. The sigmoid activation function used in the policy module is denoted as $\sigma$. The important components of the utilized DL architecture are defined in the followings.

**(1) State**

The state $s_t \in S$ is comprised of three components: the word vectors $x$ encoded by the representation module, the hidden vector $h$, and the memory vector $c$. The two latter vectors $h$ and $c$ are derived from the long short-term memory (LSTM) model of the representation module described in the next section. The state $s_t$ is updated by the equation (6).

$$s_t = x_t \oplus h_{t-1} \oplus c_{t-1}, \tag{6}$$

where $x_t$ represents the word vector of the element at the current position in the sequence. The variables $h_{t-1}$ and $c_{t-1}$ represent the hidden vector and the memory vector of the LSTM at the time scale (*t*-1). The symbol $\oplus$ denotes the horizontal concatenation of vectors.

**(2) Action**

The decision making of the policy module is to keep or discard a particular element of the represented feature vector, and is represented as a binary decision vector, with the options are either 0 (discard) or 1 (keep). Consequently, the action space is defined as a binary space, as illustrated in the equation (7).

$$a_t \in \{0, 1\}. \tag{7}$$

It is worth emphasizing that TemporalPaD employs a probabilistic approach to generate function values in the policy module during the training phase. However, during the testing phase, TemporalPaD switches to a deterministic strategy by selecting the $\pi(a_t|s_t;\theta)$ that corresponds to the largest function value, referred to as $\pi^*$, as shown in equation (8).

$$a_t = \begin{cases} \pi(a_t|s_t;\theta), & \text{Trainning step} \\ argmax_a \pi^*(a|s_t;\theta_1), & \text{Testing step} \end{cases} \tag{8}$$

**(3) Reward Function**

The reward function uses the objective of a Markov decision process (MDP) to find the trajectory (denoted as $s_1, a_1, \cdots, s_T, a_T$) that maximizes the reward. We divide the reward into two parts *A* and *B*. The reward part $A_i(k)$ has three versions and the user may choose one version manually or based on the evaluation experiment, as defined in the equation (9).

$$A_i(k) = \begin{cases} -\sum_{j=1}^{C} y_{ij} log P(\hat{y}_{ij}|X) & (k = 0) \\ \max \left(\log\left(P(\hat{y}_{ij}|X)\right)\right)_{j=1,2,\cdots,C} & (k = 1), \\ \chi(\max_j(P(\hat{y}_{ij}|X))) & (k = 2) \end{cases} \tag{9}$$

where *i* represents the $i^{th}$ sample, *C* denotes the number of categories, $y_{ij}$ and $P(\hat{y}_{ij}|X)$ respectively stand for the one-hot encoding of true labels and the predicted distribution of the sample $X$. The denotation $\chi(\hat{y}_i)$ is 1 or 0 if $\hat{y}_i$ is equal to $y_i$ or not. Here, we employ *softmax* to compute the probability distribution. $A_i(0)$ is the cross-entropy loss function, $A_i(1)$ is the logarithmic maximum value of the output neurons of the classification module, and $A_i(2)$ is the characteristic function value, respectively.

We add the regularization term $B_i$ to encourage the policy module to remove as many features as possible.

$$B = \eta \times L'/L, \tag{10}$$

where $L$ denotes the original length of a sequence, and $L'$ is the number of discarded elements, i.e., the number of $a_t=0$. The regularization parameter $\eta$ is used to balanced the two parts $A$ and $B$.

$$reward = A + B \tag{11}$$

**(4) Objective function**

This study employs the policy gradient strategy [32] to update the parameters of the policy network and to maximize the obtained reward, as described in [33]. The update process can be formulated by the equation (12).

$$J(\theta) = \mathbb{E}_{(s_t,a_t) \sim P_\theta(s_t,a_t)} r(s_1, a_1, \cdots, s_T, a_T) \tag{12}$$

$$= \sum_{s_1,a_1,\cdots,s_T,a_T} \prod_t \pi(a_t|s_t;\theta) R$$

We simplify the expression in the equation (12) by utilizing the likelihood ratio strategy [34]. Then we derive the gradient formula by the equation (13).

$$\nabla_\theta J(\theta) = \sum_{t=1}^{L} R \nabla_\theta log\pi_\theta(a_t|s_t) \tag{13}$$

$$= (\sum_{t=1}^{L} (A+B) \cdot \nabla_\theta log\pi_\theta(a_t|s_t))$$

The gradient is integrated into the equation (14) to update the parameters of the policy module [33]

$$\theta = \theta + \alpha \nabla_\theta J(\theta) \qquad (14)$$

The learning rate $\alpha$ is utilized in this study. The parameter $\theta$ is updated according to formula (14) to maximize the objective function (12), thereby enhancing the performance of downstream prediction tasks.

## Representation module

Traditional machine learning approaches heavily rely on the manually constructed features, while the emergence of deep learning revolutionizes this process by enabling the automatic mapping of raw data to a feature space through embedding techniques. The TemporalPaD framework is proposed in this study to focus on the extraction of temporal relevance between the features by the utilization of the sequential pattern extraction approach LSTM. The normalization layer in this feature extraction approach standardizes the features extracted from the input samples, resulting in the EncodeSeq dataset. This dataset is regarded as the elements constituting the state space for the policy module. The policy module iterates through the feature vectors at each position within EncodeSeq, making decisions on whether to retain the current feature. Upon completing the traversal of EncodeSeq, a decision sequence is obtained, and this decision sequence is utilized to map the original features, yielding the reduced feature set denoted as Rseq. Subsequently, Rseq is integrated into the classification module to compute the value of the reward function.

It is important to note that if at time *t*, the policy module in state $s_t$ chooses to discard the word vector $x_t$ ($a_t$=0), then $x_t$ will make no contribution to subsequent feature extraction. Conversely, if it is retained, it is utilized for further feature extraction, as depicted in Equation (15). The variable $c_t$ represents the memory unit within the LSTM, playing a pivotal role in preserving and propagating pertinent information [35]. Similarly, $h_t$ signifies the hidden vector, capturing the encoded representation of the input sequence at a specific time step [36].

$$c_t, h_t = \begin{cases} c_{t-1}, h_{t-1}, & a_t = 0 \\ \varphi(c_{t-1}, h_{t-1}, x_t), & a_t = 1 \end{cases} \qquad (15)$$

The activation function of the LSTM network is denoted as $\varphi$, which enables the LSTM network to selectively store, update, and retrieve information[37].

## Classification module

The classification module receives the masked features, Rseq, generated by the policy module, which applies the decision sequence to the feature representation EncodeSeq from the representation module. Specifically, the classification module utilizes the hidden variable $h_T$ of Rseq at the final time step as input and employs softmax as the activation function, as illustrated in Equation (16).

$$P(\widehat{y}|X) = \Psi(W_c h_T + b_c) \tag{16}$$

Here, $\Psi$ represents the *softmax* activation function. $W_c$ and $b_c$ are the network parameters within the classification module, and they are trained by the cross-entropy loss function:

$$L = \sum_{i=1}^{N} - \sum_{j=1}^{C} y_{ij} log P(\hat{y}_{ij}|X), \tag{17}$$

where, $N$ represents the sample size, $C$ denotes the number of categories, and $y_{ij}$ signifies the one-hot encoding of the true labels.

## Training process

A three-step training procedure is used to train the TemporalPaD framework from scratch, as modified from [33].

**Step 1: Pre-training**

The pre-training of the representation module and the classification module are conducted in this step. We input the samples in the original feature space to the representation module with the random initialization, and the extracted features are then passed to the classification module for classifier training. The parameters of both modules are updated by the loss function defined in the equation (17).

This step aims to learn meaningful representations from the input samples in the original feature space and accurate classification models. The pre-trained parameters serve as the initial values of the two modules for the subsequent steps.

**Step 2: Reinforcement Learning**

This step fixes the parameters of the representation module and the classification module, and optimizes the decision-making capability of the policy module. The sequence of the original features is subjected to processing by the policy module, resulting in the generation of a binary vector, where each element specifies whether a feature is kept (1) or discarded (0). Then, the original feature sequence is mapped to the feature-selected sequence denoted as Rseq using this binary vector and is subsequently supplied to the classification module for reward computation. The parameters of the policy module are updated by the equations (13) and (14).

**Step 3: Fine-tuning**

This step jointly trains the three modules of the TemporalPaD framework. The joint training aims to optimize their collaborations in the feature representation task.

# Results and discussion

## Evaluation of the Structured Dataset

Although TemporalPaD is an end-to-end framework specifically designed for feature extraction and dimensionality reduction of time series data, it can still be effectively compared with various feature dimensionality reduction algorithms on structured datasets. These datasets serve as comprehensive benchmarks for evaluating feature reduction algorithms without altering the model structure. Before presenting the experiments, we provide a detailed description of the model structure, reward function, and voting strategy used to ensure a consistent feature subset across all samples of the structured data when evaluating TemporalPaD on the UCI dataset.

The policy module of TemporalPaD is implemented using a two-layer fully connected neural network, aligning with the principles of reinforcement learning, which prioritize simpler network architectures [38]. The representation module employs embedding techniques to extract high-dimensional features from the UCI datasets, followed by a single-layer LSTM network to capture temporal correlations among these features. The classification module then utilizes a single-layer fully connected neural network with a softmax activation function to generate probability distributions. The reward function is computed according to the method described as A(1) in formula (9).

To maintain consistent feature numbers for each sample post-dimension reduction, we incorporate a voting mechanism. A threshold is set, and TemporalPaD eliminates a feature from all samples only when the proportion of its removal, as determined by the policy module, exceeds this threshold. All subsequent experiments on the UCI dataset utilize this consistent model configuration.

## 10 Independent Tests on 29 UCI datasets

We first conduct a comparative evaluation of TemporalPaD, an algorithm for feature extraction based on temporal dependencies, with five feature selection algorithms based on feature correlations, using 29 UCI datasets in 10 independent validation tests. Each test employs a different random seed to ensure reliable results. Performance metrics are assessed by calculating the average accuracy (Acc) and its corresponding standard deviation for each method. The average Acc from the 10 independent tests provides an overall measure of the model's accuracy, while the standard deviation indicates the variability in the results. Following a methodology similar to [39], we examine how often different algorithms achieve the highest average Acc across all datasets.

When using Support Vector Machine (SVM) as the downstream classifier, TemporalPaD demonstrates exceptional performance, achieving the highest average Acc on 16 out of the 29 datasets (refer to Table 2). Furthermore, when Naive Bayes (NB) is used as the downstream classifier, TemporalPaD outperforms other methods and attains the highest average Acc on 8 datasets (refer to Table 4). These results demonstrate that, compared to feature selection techniques based on feature correlations, the TemporalPaD approach for feature extraction and dimension reduction, employing Deep Reinforcement Learning (DRL), exhibits superior performance within this framework. It is also important to note that the comparison reveals the relatively lower effectiveness of TemporalPaD with the C4.5 classifier (refer to Table 3). This finding emphasizes the significance of carefully selecting a compatible downstream classifier when utilizing a feature reduction method [40].

**Table 2:** Average Acc and standard deviation from 10 independent tests conducted on TemporalPaD and five feature selection algorithms based on feature relevance, using SVM as the downstream classifier. Values in parentheses indicate the corresponding standard deviations.

| | SVM | | | | | | |
|---|---|---|---|---|---|---|---|
| Data | TemporalPaD(Ours) | Full set | CFS | Consistency | FCBF | INTERACT | MRMI |
| Arrhythmia | **64.78(9.50)** | 53.83 | 53.83(3.36) | 53.83(3.36) | 54.16(3.29) | 53.83(3.36) | 53.58(3.27) |
| Austra | **79.61(12.27)** | 54.09 | 56.85(3.24) | 54.04(2.58) | 56.85(3.24) | 54.0(2.53) | 70.55(1.85) |
| Breastcancer | 94.14(4.44) | 96.13 | 96.13(1.21) | **96.30(1.03)** | 95.80(1.19) | **96.30(1.03)** | 95.80(1.19) |
| Chess | 89.80(12.01) | 97.34 | 94.31(0.46) | **98.11(0.41)** | 94.31(0.46) | 97.71(0.41) | 94.36(0.46) |
| Credit | **84.78(4.85)** | 56.81 | 56.34(2.04) | 56.72(2.20) | 56.34(2.04) | 56.72(2.46) | 58.21(3.26) |
| Diabetes | **69.87(6.06)** | 64.83 | 63.14(1.92) | 64.83(2.35) | 63.14(1.92) | 64.83(2.35) | 63.14(1.92) |
| German | 71.20(3.45) | 69.21 | **71.38(2.31)** | 69.21(2.13) | **71.38(2.31)** | 69.21(2.13) | **71.38(2.31)** |
| Glass | 61.56(8.81) | 64.11 | 63.84(5.82) | 65.48(4.69) | 64.25(4.06) | 65.48(4.69) | **68.36(3.90)** |
| Heart | **76.91(13.10)** | 55.65 | 74.78(3.79) | 64.35(4.03) | 74.78(3.79) | 64.35(4.03) | 75.11(4.65) |
| Horse colic | 75.82(6.64) | 66.48 | 76.72(3.23) | 65.20(4.77) | 76.72(3.23) | 67.20(3.27) | 70.40(2.20) |
| Hypothyroid | **97.15(1.17)** | 95.09 | 95.30(0.62) | 95.19(0.62) | 95.30(0.62) | 95.19(0.62) | 95.70(0.57) |
| Ionosphere | 89.81(2.66) | 92.27 | **94.37(1.90)** | 90.17(2.69) | 88.99(2.81) | 91.93(2.14) | 90.50(2.21) |
| Iris | 92.89(8.63) | **96.86** | 96.27(0.35) | 96.27(0.35) | 96.27(0.35) | 96.27(0.35) | 96.27(0.35) |
| Liver Disorders | **61.63(5.26)** | 58.89 | 58.89(4.14) | 58.89(4.14) | 58.89(4.14) | 58.89(4.14) | 58.89(4.14) |
| Lung cancer | 42.00(14.76) | 32.73 | 58.18(9.77) | 66.36(16.06) | 55.45(19.85) | **70.00(9.63)** | 58.18(9.77) |
| Mushroom | 93.38(10.64) | **100** | 98.95(0.18) | 99.85(0.12) | 98.95(0.18) | **100.00(0.00)** | 98.66(0.10) |
| Pima | **72.22(6.56)** | 65.25 | 65.25(1.36) | 65.25(1.36) | 65.25(1.36) | 65.25(1.36) | 65.25(1.36) |
| Segmentation | **81.15(15.78)** | 54.62 | 48.56(4.39) | 49.72(3.96) | 45.13(3.42) | 46.47(4.05) | 80.55(1.71) |
| Sonar | **75.81(10.78)** | 56.47 | 69.72(7.21) | 72.25(5.98) | 66.48(7.21) | 69.15(7.76) | 63.80(3.52) |
| Spambase | 87.14(9.32) | 82.47 | **91.78(0.71)** | 80.86(1.05) | 90.18(0.65) | 81.16(1.30) | 89.59(0.93) |
| Spect | **80.50(5.81)** | 71.76 | 72.42(4.93) | 70.88(4.40) | 72.97(5.13) | 70.88(4.40) | 72.98(5.41) |
| Vehicle | **63.07(10.92)** | 53.58 | 45.49(3.44) | 53.58(4.43) | 40.56(2.75) | 53.58(4.43) | 38.33(2.46) |
| Vote | 89.92(8.62) | 95.34 | **95.74(1.28)** | 95.68(1.20) | **95.74(1.28)** | 95.27(1.27) | **95.74(1.28)** |
| Waveform | 80.82(17.29) | 86.5 | **86.74(0.61)** | 82.89(0.59) | 78.74(0.60) | 85.09(0.70) | 78.05(1.53) |
| WDBC | 90.99(12.34) | 62.38 | 62.75(3.36) | 93.88(0.77) | **94.51(1.66)** | 71.30(4.97) | 93.21(1.57) |
| Wine | **95.66(6.10)** | 41.8 | 40.66(3.07) | 92.62(2.35) | 40.66(3.07) | 49.34(4.47) | 93.61(1.80) |
| WPBC | **78.14(4.26)** | 74.33 | 72.39(2.36) | 72.39(2.36) | 72.39(2.36) | 72.39(2.36) | 71.04(1.75) |
| Yeast | **57.57(3.36)** | 40.44 | 41.15(1.56) | 41.15(1.56) | 41.47(1.46) | 41.15(1.56) | 45.74(1.30) |
| Zoo | **85.00(15.42)** | 83.53 | 82.35(11.0) | 82.65(6.86) | 81.76(8.30) | 82.65(6.86) | 77.94(7.99) |
| NumBest | 16 | 2 | 6 | 2 | 4 | 3 | 3 |

**Table 3:** Average Acc and standard deviation from 10 independent tests conducted on TemporalPaD and five feature selection algorithms based on feature relevance, using C4.5 as the downstream classifier. Values in parentheses indicate the corresponding standard deviations.

| | C4.5 | | | | | | |
|---|---|---|---|---|---|---|---|
| Data | TemporalPaD(Ours) | Full set | CFS | Consistency | FCBF | INTERACT | MRMI |
| Arrhythmia | 46.99(7.06) | 63.51 | 69.13(4.03) | 69.72(4.33) | **71.04(4.51)** | 66.88(2.77) | 63.27(2.51) |
| Austra | 74.69(9.17) | 83.91 | 83.32(2.51) | 83.83(3.39) | 83.49(1.85) | 83.75(2.30) | **84.73(1.88)** |
| Breastcancer | 93.00(3.73) | 94.87 | 94.87(1.69) | 94.71(1.47) | **95.38(0.96)** | 95.71(1.52) | 95.13(1.29) |
| Chess | 90.36(13.02) | 97.97 | 94.31(0.49) | **99.03(0.46)** | 94.31(0.49) | 98.77(0.49) | 94.36(0.46) |
| Credit | 80.58(5.14) | 85.83 | 85.28(1.02) | 84.68(1.08) | 84.77(1.50) | 85.65(1.47) | **85.66(1.32)** |
| Diabetes | 67.13(5.82) | **74.77** | 73.95(2.38) | **74.77(2.13)** | 73.74(2.18) | 73.95(2.38) | 73.87(2.49) |
| German | 66.03(3.18) | 70.26 | 70.47(0.57) | 71.26(2.06) | 71.32(1.52) | 72.10(2.28) | **73.38(1.61)** |
| Glass | 58.59(9.44) | **68.36** | 66.87(2.78) | 64.13(5.97) | 68.09(5.17) | 64.79(4.18) | 66.71(4.14) |
| Heart | 74.32(6.32) | 75 | 76.09(3.69) | 74.13(2.36) | 77.42(4.67) | 76.69(2.67) | **78.58(4.53)** |
| Horse colic | 67.45(7.96) | 65.79 | **70.70(3.35)** | 67.04(3.12) | 68.49(4.49) | 67.44(3.40) | 68.40(3.31) |
| Hypothyroid | 97.50(1.37) | 99.3 | 99.30(0.21) | **99.31(0.17)** | 99.30(0.21) | 99.10(0.28) | 98.16(0.33) |
| Ionosphere | 89.52(2.01) | 89.08 | 88.74(4.43) | 88.24(2.66) | **90.08(1.54)** | 90.01(3.72) | 88.07(3.62) |
| Iris | 92.67(7.26) | 95.1 | 95.29(0.48) | 95.29(0.48) | 95.29(0.48) | 95.29(0.48) | 95.29(0.48) |
| Liver Disorders | 59.90(5.93) | **61.45** | 59.23(2.49) | 59.23(2.49) | 59.23(2.49) | 59.23(2.49) | 59.23(2.49) |
| Lung cancer | 42.00(9.19) | 43.64 | 50(18.80) | 54.55(17.14) | 45.45(10.03) | 55.45(20.31) | **59.09(15.88)** |
| Mushroom | 93.47(10.67) | **100** | 99.02(0.12) | **100.00(0.00)** | 99.02(0.12) | **100.00(0.00)** | 98.58(0.13) |
| Pima | 67.13(6.37) | 72.48 | 73.72(3.98) | 72.48(3.62) | 66.87(2.83) | 73.10(3.07) | **73.79(2.91)** |
| Segmentation | 84.63(15.21) | 95.96 | **96.09(0.86)** | 95.96(0.90) | 94.92(0.74) | 95.80(0.73) | 95.53(0.81) |
| Sonar | 66.94(11.44) | 70.42 | 69.58(7.49) | **76.34(10.37)** | 69.58(6.37) | 71.55(5.13) | 70.00(3.64) |
| Spambase | 85.49(8.04) | 92.22 | 92.18(0.72) | 91.21(0.65) | **92.62(0.48)** | 91.68(0.65) | 91.47(0.68) |
| Spect | **75.75(5.01)** | 69.23 | 72.42(5.32) | 70.88(5.11) | 71.87(5.13) | 70.88(5.11) | 71.87(5.13) |
| Vehicle | 62.13(9.54) | **70.59** | 66.72(2.35) | 70.59(1.53) | 56.28(2.21) | 70.59(1.53) | 59.34(2.55) |
| Vote | 87.94(8.20) | **95.95** | 95.41(0.77) | 95.95(0.91) | 95.41(0.77) | 95.68(1.07) | 95.57(0.90) |
| Waveform | 69.07(12.78) | 74.86 | **76.81(1.10)** | 75.91(0.97) | 75.15(0.98) | 76.38(0.59) | 74.56(0.80) |
| WDBC | 86.67(16.22) | 93.96 | 94.30(1.47) | 94.30(1.94) | 93.78(1.72) | 94.24(1.69) | **94.55(1.13)** |
| Wine | 92.08(5.32) | 90.16 | 92.46(3.59) | 94.10(2.74) | 90.33(7.39) | **94.67(3.25)** | 91.15(5.48) |
| WPBC | 66.78(7.80) | 71.04 | 73.58(3.73) | 73.58(3.73) | 73.58(3.73) | 73.58(3.73) | **74.63(2.33)** |
| Yeast | 50.02(4.03) | 52.97 | 53.56(4.47) | 53.56(4.47) | **53.62(1.94)** | 53.56(4.47) | 53.01(3.74) |
| Zoo | 84.67(15.49) | **93.24** | 91.76(4.96) | 88.82(6.01) | 89.71(5.92) | 89.41(6.23) | 89.71(5.92) |
| NumBest | 1 | 8 | 4 | 8 | 5 | 5 | 8 |

**Table 4:** Average Acc and standard deviation from 10 independent tests conducted on TemporalPaD and five feature selection baseline algorithms based on feature relevance, using NB as the downstream classifier. Values in parentheses represent the corresponding standard deviations.

| | NB | | | | | | |
|---|---|---|---|---|---|---|---|
| Data | TemporalPaD(Ours) | Full set | CFS | Consistency | FCBF | INTERACT | MRMI |
| Arrhythmia | 62.13(9.01) | 60.45 | **70.43(3.82)** | 69.24(4.34) | 68.90(2.36) | 68.47(3.84) | 66.33(3.99) |
| Austra | **79.37(11.58)** | 76.09 | 75.32(3.24) | 74.89(3.25) | 73.57(2.11) | 74.77(2.04) | 74.33(3.48) |
| Breastcancer | 93.10(5.56) | 96.72 | 96.72(2.03) | 95.88(0.69) | **96.76(1.05)** | 96.64(1.19) | 96.68(1.13) |
| Chess | 80.79(8.25) | 86.62 | 92.64(0.79) | 88.90(1.30) | 92.64(0.79) | 89.24(1.80) | **94.36(0.46)** |
| Credit | **83.67(4.56)** | 78.13 | 74.81(1.89) | 75.32(2.57) | 75.32(2.27) | 76.53(1.58) | 74.81(2.32) |
| Diabetes | 69.04(7.13) | 75.16 | 77.05(2.41) | 75.16(2.04) | 76.44(2.23) | 77.05(2.41) | **77.21(2.48)** |
| German | 70.23(2.06) | **74.32** | 73.24(2.21) | 73.87(2.17) | 73.76(2.31) | 74.01(2.39) | 73.88(2.18) |
| Glass | **54.84(9.06)** | 48.22 | 46.45(5.36) | 46.82(6.21) | 46.16(4.14) | 47.26(3.24) | 47.40(6.37) |
| Heart | 76.79(13.33) | **84.67** | 83.70(2.17) | 82.61(2.17) | 84.24(1.87) | 83.45(1.94) | 83.80(2.78) |
| Horse colic | **74.73(8.59)** | 65.32 | 65.12(2.86) | 64.16(2.22) | 64.76(2.12) | 66.16(3.68) | 64.56(2.07) |
| Hypothyroid | 95.43(0.64) | 97.57 | 97.66(0.27) | **97.89(0.27)** | 97.66(0.27) | 97.62(0.41) | 97.44(0.40) |
| Ionosphere | 80.29(9.50) | 82.5 | **89.41(3.97)** | 85.88(5.13) | 85.17(3.42) | 86.93(3.18) | 87.48(2.45) |
| Iris | 92.22(9.38) | **96.67** | 96.67(2.78) | 96.67(2.78) | 96.67(2.78) | 96.67(2.78) | 96.67(2.78) |
| Liver Disorders | **60.48(4.07)** | 54.27 | 56.07(2.94) | 56.07(2.94) | 56.07(2.94) | 56.07(2.94) | 56.07(2.94) |
| Lung cancer | 44.00(14.30) | 44.55 | 70.00(8.62) | 61.82(15.92) | 55.45(13.85) | 70.91(12.71) | **71.82(9.49)** |
| Mushroom | 86.38(8.05) | 95.29 | 98.58(0.13) | 98.55(0.29) | 98.58(0.13) | **98.89(0.17)** | 98.58(0.13) |
| Pima | 71.30(5.60) | 74.31 | **76.02(2.39)** | 74.31(2.89) | 65.34(3.04) | 74.25(2.16) | 75.02(2.23) |
| Segmentation | 76.00(16.23) | 80.36 | 86.89(0.86) | 79.50(1.50) | **87.13(1.27)** | 79.63(0.98) | 83.43(1.68) |
| Sonar | 71.29(12.14) | 70.28 | 70.70(6.56) | 68.45(3.93) | **73.66(3.32)** | 70.56(7.87) | 70.99(5.24) |
| Spambase | 79.11(13.03) | 79.93 | 78.80(0.68) | 86.96(1.05) | 77.10(0.86) | **88.64(0.81)** | 85.29(3.55) |
| Spect | **77.50(4.75)** | 69.23 | 68.90(2.93) | 72.86(3.52) | 71.87(5.13) | 72.86(3.52) | 71.87(5.13) |
| Vehicle | **51.54(10.14)** | 45.66 | 47.72(3.80) | 45.66(3.09) | 40.49(2.21) | 45.66(3.09) | 41.67(2.28) |
| Vote | 88.85(8.49) | 90.54 | 95.68(0.47) | 91.28(1.67) | 95.68(0.47) | 92.97(1.60) | **96.32(0.96)** |
| Waveform | 76.50(15.65) | 80.13 | 80.19(0.87) | 81.14(0.75) | 77.90(0.80) | **81.76(0.88)** | 77.12(0.85) |
| WDBC | 88.13(14.28) | **96.61** | 94.53(1.01) | 94.36(1.41) | 95.45(0.89) | 94.76(1.05) | 95.79(1.17) |
| Wine | 87.17(14.66) | 97.7 | 97.70(2.49) | 95.74(1.87) | **98.20(1.80)** | 96.72(1.52) | 96.07(2.81) |
| WPBC | **77.97(4.15)** | 64.33 | 74.63(3.15) | 74.63(3.15) | 74.63(3.15) | 74.63(3.15) | 72.69(3.23) |
| Yeast | 45.48(4.30) | 55.17 | **56.06(1.90)** | 56.06(1.90) | 54.57(2.34) | 56.06(1.90) | 54.02(2.54) |
| Zoo | 79.33(14.56) | **94.71** | 92.65(4.44) | 86.76(6.54) | 90.29(6.51) | 86.76(6.54) | 88.24(6.93) |
| NumBest | 8 | 5 | 5 | 3 | 5 | 5 | 5 |

# 10-fold Cross-Validation with the Same Features as TemporalPaD on 10 UCI Datasets

We conduct a 10-fold cross-validation experiment based on the UCI dataset to further verify the performance of TemporalPaD on structured data [41]. We select a total of 10 binary datasets from the subject areas "health and medicine" listed in Table 1, referred to as UCIH&M. For the comparisons, we consider 18 dimensionality reduction methods, comprising 8 information-theory-based feature selection methods and 10 feature extraction algorithms. We ensure that each method select the same number of features as TemporalPaD in each fold, allowing for a fair comparison of the performance of different methods with an equal number of features.

The analysis of Tables 5 to 7 reveals that the "NumBest" values indicate a general superiority of feature selection algorithms over feature extraction algorithms on the

UCIH&M dataset. This observation can be attributed to the structured nature of UCIH&M. Feature selection algorithms maintain the original meaningful representation, preserving interpretability and the inherent meaning of features without alteration [13]. In contrast, feature extraction techniques transform the original features into deeper abstract features through mapping.

A closer evaluation focusing on feature extraction shows that TemporalPaD exhibits satisfactory results when compared to the other 10 feature extraction methods. Notably, TemporalPaD achieves the highest average Acc on the Liver Disorders dataset when used in conjunction with SVM and C4.5 classifiers (refer to Tables 5 and 7). It is worth mentioning that TemporalPaD is a more comprehensive feature extraction method, applicable to various dataset types and not constrained by the rank of the covariance matrix of the dataset. This characteristic explains why PCA, LLE, and UMAP methods fail to extract features from the Liver Disorders dataset while maintaining the same number of selected features as TemporalPaD [42-44]. In other words, in certain folds, TemporalPaD selects a larger number of features than the sample size, resulting in the failure of PCA, LLE, and UMAP.

**Table 5:** The 10-fold cross-validation evaluation of 8 feature selection algorithms grounded in information theory, alongside 10 feature extraction methods on 10 UCI datasets. Each algorithm selects the same number of features as TemporalPaD in each fold of the 10-fold cross-validation. This evaluation is performed with support vector machines (SVM) serving as the classifier. The highest average Acc attained for each dataset is denoted in bold.

| SVM | Breast | Diabetes | Heart | hypothyroid | Liver Disorders | Lung-cancer | Pima | spect | wdbc | WPBC | NumBest |
|---|---|---|---|---|---|---|---|---|---|---|---|
| Feature Extraction | | | | | | | | | | | |
| TemporalPaD (Ours) | 0.9585(0.0334) | 0.7565(0.0717) | 0.7111(0.15) | 0.9693(0.0129) | **0.6701(0.0793)** | 0.4333(0.2881) | 0.7384(0.0491) | 0.8125(0.0473) | 0.9122(0.1207) | 0.7887(0.0863) | 1 |
| FA | 0.9684(0.0225) | 0.7421(0.0504) | 0.8407(0.1077) | 0.9728(0.0131) | 0.6437(0.0741) | 0.425(0.2734) | 0.7381(0.0661) | 0.8312(0.0651) | 0.9719(0.0323) | 0.7634(0.0987) | 0 |
| PCA | **0.9728(0.0173)** | 0.7708(0.0599) | 0.8333(0.0586) | 0.9829(0.0096) | 0.6205(0.0532) | - | 0.7694(0.0591) | 0.8352(0.0535) | 0.9666(0.024) | 0.7932(0.0718) | 1 |
| KernelPCA | 0.9699(0.0198) | 0.7734(0.0649) | 0.8333(0.0586) | 0.9839(0.0096) | 0.6176(0.05) | 0.4583(0.2332) | 0.7668(0.0601) | 0.8315(0.0562) | 0.9684(0.0216) | 0.7982(0.074) | 0 |
| SparsePCA | 0.9713(0.0204) | 0.7695(0.0589) | 0.8333(0.0636) | 0.9829(0.0096) | 0.6234(0.059) | 0.4917(0.2372) | 0.7694(0.0591) | 0.8315(0.0535) | 0.9684(0.0259) | 0.7982(0.0777) | 0 |
| SVD | 0.9713(0.0167) | 0.7747(0.0543) | 0.8296(0.0765) | 0.9772(0.0088) | 0.6384(0.0748) | 0.4667(0.2699) | 0.7733(0.0479) | 0.8387(0.0565) | 0.9666(0.021) | 0.7634(0.0987) | 0 |
| ICA | 0.9599(0.015) | 0.767(0.0575) | 0.8148(0.0605) | 0.9813(0.0081) | 0.6118(0.068) | 0.3667(0.1721) | 0.7681(0.0589) | 0.8426(0.0554) | 0.9631(0.0374) | 0.7784(0.1074) | 0 |
| NMF | 0.9685(0.0177) | 0.7631(0.0434) | 0.8111(0.075) | 0.9495(0.0742) | 0.6092(0.0782) | 0.4917(0.2436) | 0.7734(0.0522) | 0.839(0.0524) | 0.9613(0.0284) | 0.7787(0.1073) | 0 |
| ISOMAP | 0.9713(0.0152) | 0.7421(0.0507) | 0.8259(0.0654) | 0.964(0.0062) | 0.6324(0.0741) | **0.525(0.1759)** | 0.7421(0.0507) | 0.7789(0.0619) | 0.9613(0.0259) | 0.7582(0.0954) | 1 |
| LLE | 0.9427(0.0735) | 0.7148(0.0564) | 0.8074(0.0887) | 0.9681(0.01) | 0.6266(0.0674) | - | 0.6966(0.0711) | **0.8538(0.0541)** | 0.9648(0.0186) | 0.7632(0.1097) | 1 |
| UMAP | 0.967(0.0216) | 0.737(0.0306) | 0.8074(0.0834) | 0.9542(0.0045) | 0.6324(0.0741) | - | 0.7303(0.0661) | 0.7521(0.0996) | 0.9473(0.0233) | 0.7634(0.0987) | 0 |
| Feature Selection | | | | | | | | | | | |
| CIFE | 0.947(0.0227) | 0.7734(0.0454) | 0.8148(0.08) | **0.9845(0.0059)** | 0.647(0.0655) | 0.4917(0.2845) | 0.7734(0.0454) | 0.816(0.0852) | 0.7838(0.1319) | 0.7634(0.0987) | 1 |
| CMIM | **0.9728(0.0173)** | 0.7747(0.0503) | **0.8481(0.077)** | 0.9823(0.0064) | 0.647(0.0655) | **0.525(0.2834)** | 0.7747(0.0503) | 0.8275(0.0539) | **0.9736(0.0223)** | 0.7787(0.1099) | 4 |
| DISR | 0.9713(0.0192) | 0.7812(0.0464) | **0.8481(0.0686)** | 0.9807(0.0066) | 0.647(0.0655) | **0.525(0.2834)** | 0.7747(0.0503) | 0.8275(0.0539) | 0.9719(0.0222) | **0.8182(0.0716)** | 3 |
| ICAP | **0.9728(0.0173)** | 0.7747(0.0503) | **0.8481(0.077)** | 0.9823(0.0064) | 0.647(0.0655) | **0.525(0.2834)** | 0.7747(0.0503) | 0.8275(0.0539) | **0.9736(0.0223)** | 0.7787(0.1099) | 4 |
| JMI | **0.9728(0.0173)** | 0.7799(0.0462) | **0.8481(0.077)** | 0.9823(0.0069) | 0.647(0.0655) | **0.525(0.2834)** | 0.7799(0.0462) | 0.8313(0.0538) | **0.9736(0.0223)** | **0.8182(0.0676)** | 5 |
| MIFS | 0.9484(0.0249) | **0.785(0.0513)** | 0.7185(0.0765) | **0.9845(0.0072)** | 0.6092(0.0702) | 0.4167(0.2833) | **0.785(0.0513)** | 0.812(0.0948) | 0.9051(0.0381) | 0.7634(0.0987) | 3 |
| MIM | **0.9728(0.0173)** | 0.7799(0.0462) | **0.8481(0.077)** | 0.9823(0.0069) | 0.647(0.0655) | **0.525(0.2834)** | 0.7799(0.0462) | 0.8313(0.0538) | **0.9736(0.0223)** | **0.8182(0.0676)** | 5 |
| MRMR | **0.9728(0.0173)** | 0.7799(0.0555) | 0.6259(0.1177) | 0.9839(0.0064) | 0.6092(0.0702) | 0.4917(0.2845) | 0.7799(0.0555) | 0.8503(0.0491) | 0.7049(0.0629) | 0.7634(0.0987) | 1 |

**Table 6:** The 10-fold cross-validation evaluation of 8 feature selection algorithms

grounded in information theory, alongside 10 feature extraction methods on 10 UCI datasets, employing Naive Bayes (NB) as the classifier, with each algorithm selecting the same number of features as TemporalPaD in each fold of the 10-fold cross-validation. The highest average Acc attained for each dataset is denoted in bold.

| NB | Breast | horse-colic | Heart | hypothyroid | Liver Disorders | Lung-cancer | Pima | spect | wdbc | WPBC | NumBest |
|---|---|---|---|---|---|---|---|---|---|---|---|
| Feature Extraction | | | | | | | | | | | |
| TemporalPaD (Ours) | 0.9527(0.0318) | 0.7513(0.0676) | 0.7259(0.1494) | 0.6276(0.2581) | 0.6232(0.0869) | 0.4(0.2509) | 0.7084(0.0577) | 0.7641(0.0718) | 0.8894(0.1207) | 0.7016(0.1054) | 0 |
| FA | 0.9684(0.0326) | 0.7487(0.0485) | 0.8185(0.075) | 0.7612(0.308) | 0.5975(0.097) | 0.4917(0.1776) | 0.742(0.0816) | 0.8202(0.0652) | 0.9525(0.0332) | 0.7132(0.0859) | 0 |
| PCA | 0.957(0.0236) | 0.7734(0.0565) | 0.8296(0.0745) | **0.9734(0.0148)** | 0.6005(0.1052) | - | 0.7694(0.0625) | 0.8201(0.0634) | 0.942(0.0275) | **0.7742(0.1243)** | 2 |
| KernelPCA | 0.9628(0.0141) | 0.7617(0.0505) | 0.8259(0.0699) | 0.9703(0.0143) | 0.6035(0.105) | **0.5833(0.2357)** | 0.7746(0.056) | **0.8313(0.0594)** | 0.949(0.0229) | 0.7589(0.1136) | 2 |
| SparsePCA | 0.9613(0.0217) | 0.7604(0.0597) | 0.8296(0.0841) | 0.8749(0.1672) | 0.6211(0.094) | 0.5333(0.2194) | 0.7564(0.0563) | 0.7903(0.0897) | 0.9508(0.0216) | 0.7376(0.0965) | 0 |
| SVD | 0.9585(0.0209) | 0.7643(0.0387) | 0.8444(0.0777) | 0.9374(0.0859) | 0.6266(0.0725) | 0.5(0.1571) | 0.75(0.0409) | 0.8201(0.058) | 0.9473(0.0165) | 0.7434(0.1078) | 0 |
| ICA | 0.9385(0.0447) | 0.7669(0.0614) | 0.8222(0.0852) | 0.9362(0.0762) | 0.618(0.0929) | 0.3333(0.2079) | 0.7759(0.064) | 0.8202(0.0719) | 0.8909(0.0597) | 0.7129(0.15) | 0 |
| NMF | 0.9613(0.0217) | 0.767(0.0532) | 0.8111(0.077) | 0.9456(0.022) | 0.6382(0.0779) | 0.4083(0.2168) | 0.7669(0.0557) | 0.7678(0.0941) | 0.9473(0.036) | 0.7432(0.0879) | 0 |
| ISOMAP | **0.9699(0.0186)** | 0.7277(0.0478) | 0.8333(0.0766) | 0.9507(0.0119) | 0.6324(0.0741) | 0.5583(0.2292) | 0.7277(0.0478) | 0.7863(0.0752) | **0.9578(0.0169)** | 0.7126(0.1237) | 2 |
| LLE | 0.897(0.0846) | 0.6432(0.0724) | 0.8037(0.0856) | 0.8909(0.1226) | 0.5308(0.1106) | - | 0.6355(0.0735) | 0.7939(0.0848) | 0.9473(0.0233) | 0.6218(0.0862) | 0 |
| UMAP | 0.9685(0.0191) | 0.694(0.0646) | 0.8222(0.0969) | 0.9361(0.0234) | 0.6324(0.0741) | - | 0.6966(0.0515) | 0.7819(0.0982) | 0.9456(0.0153) | 0.6218(0.0862) | 0 |
| Feature Selection | | | | | | | | | | | |
| CIFE | 0.9427(0.0255) | 0.7473(0.0594) | 0.8074(0.0796) | 0.5647(0.0234) | **0.6442(0.0889)** | 0.5(0.2722) | 0.7473(0.0594) | 0.5615(0.1551) | 0.7697(0.1334) | 0.7642(0.1095) | 1 |
| CMIM | 0.9599(0.019) | 0.7604(0.0439) | **0.8556(0.0729)** | 0.5545(0.023) | **0.6442(0.0889)** | 0.4333(0.2108) | 0.7604(0.0439) | 0.5728(0.0871) | 0.9508(0.0231) | 0.7329(0.1071) | 2 |
| DISR | 0.9613(0.0168) | 0.7734(0.0547) | 0.8481(0.0845) | 0.5479(0.0225) | **0.6442(0.0889)** | 0.4333(0.2108) | 0.7734(0.0547) | 0.5654(0.0917) | 0.949(0.0241) | 0.7074(0.1012) | 1 |
| ICAP | 0.9599(0.019) | 0.7604(0.0439) | **0.8556(0.0729)** | 0.5545(0.023) | **0.6442(0.0889)** | 0.4333(0.2108) | 0.7604(0.0439) | 0.5728(0.0871) | 0.9508(0.0231) | 0.7329(0.1071) | 2 |
| JMI | 0.9599(0.019) | 0.7604(0.0439) | **0.8556(0.0729)** | 0.5454(0.0238) | **0.6442(0.0889)** | 0.4333(0.2108) | 0.7604(0.0439) | 0.5654(0.0917) | 0.949(0.0241) | 0.7132(0.1008) | 2 |
| MIFS | 0.9355(0.0284) | **0.7838(0.0466)** | 0.7148(0.0721) | 0.6267(0.1727) | 0.6181(0.089) | 0.5(0.1571) | **0.7838(0.0466)** | 0.4268(0.0985) | 0.9156(0.0528) | 0.7234(0.0985) | 2 |
| MIM | 0.9599(0.019) | 0.7734(0.0547) | **0.8556(0.0729)** | 0.546(0.0239) | **0.6442(0.0889)** | 0.4333(0.2108) | 0.7734(0.0547) | 0.5654(0.0917) | 0.949(0.0241) | 0.7332(0.1012) | 2 |
| MRMR | 0.9599(0.019) | 0.7773(0.0551) | 0.6333(0.104) | 0.5834(0.1551) | 0.6181(0.089) | 0.4333(0.2108) | 0.7773(0.0551) | 0.4756(0.0992) | 0.7014(0.0646) | 0.7234(0.0985) | 0 |

**Table 7:** The 10-fold cross-validation evaluation of 8 feature selection algorithms grounded in information theory, alongside 10 feature extraction methods on 10 UCI datasets. The employed classifier is the C4.5 decision tree, and an equivalent number of features are selected as TemporalPaD in each fold of the 10-fold cross-validation. The highest average Acc for each dataset is highlighted in bold.

| C4.5 | Breast | Diabetes | Heart | hypothyroid | Liver Disorders | Lung-cancer | Pima | spect | wdbc | WPBC | NumBest |
|---|---|---|---|---|---|---|---|---|---|---|---|
| Feature Extraction | | | | | | | | | | | |
| TemporalPaD (Ours) | 0.9312(0.0372) | 0.6981(0.0725) | 0.6556(0.1119) | 0.9728(0.0141) | **0.6522(0.0751)** | 0.5083(0.2845) | 0.7085(0.0498) | 0.7785(0.0691) | 0.8824(0.1595) | 0.6971(0.1101) | 1 |
| FA | 0.9527(0.0346) | 0.6692(0.0477) | 0.8111(0.077) | 0.9757(0.0104) | 0.5654(0.0825) | 0.4583(0.2332) | 0.6822(0.0528) | 0.7564(0.0752) | 0.9332(0.0162) | 0.6224(0.1335) | 0 |
| PCA | 0.947(0.0153) | 0.7071(0.0564) | 0.7889(0.0856) | 0.9826(0.0072) | 0.5945(0.0826) | - | 0.7032(0.0569) | 0.7637(0.0692) | 0.9455(0.0267) | 0.7284(0.1066) | 0 |
| KernelPCA | 0.9442(0.0171) | 0.7162(0.0589) | **0.8407(0.0742)** | 0.982(0.008) | 0.5678(0.1049) | 0.3917(0.3044) | 0.7187(0.0606) | 0.7712(0.0683) | 0.9455(0.0267) | 0.6932(0.1324) | 1 |
| SparsePCA | 0.9413(0.0277) | 0.7136(0.0539) | 0.7889(0.0495) | 0.9848(0.0055) | 0.5941(0.0637) | 0.4(0.2772) | 0.7071(0.0492) | 0.7785(0.0665) | 0.9473(0.0231) | 0.6726(0.109) | 0 |
| SVD | 0.9485(0.0167) | 0.7044(0.0367) | 0.7815(0.0863) | 0.9839(0.0062) | 0.5978(0.0771) | 0.4917(0.361) | 0.6952(0.0548) | 0.7632(0.0856) | 0.9368(0.0288) | 0.7276(0.0996) | 0 |
| ICA | 0.947(0.0385) | 0.7135(0.0281) | 0.7778(0.0494) | 0.9832(0.0056) | 0.5914(0.086) | 0.3833(0.3338) | 0.6966(0.0353) | 0.7561(0.0721) | 0.8787(0.0583) | **0.7382(0.0634)** | 1 |
| NMF | 0.9556(0.023) | 0.7148(0.0531) | 0.7778(0.0676) | 0.9697(0.0265) | 0.6144(0.0806) | 0.4083(0.3104) | 0.6875(0.059) | 0.7631(0.0787) | 0.9262(0.0411) | 0.7087(0.1353) | 0 |
| ISOMAP | **0.9571(0.0118)** | 0.6927(0.0311) | 0.7481(0.0869) | 0.9753(0.0055) | 0.5428(0.1011) | 0.375(0.2523) | 0.6901(0.0248) | 0.749(0.0586) | 0.9297(0.0263) | 0.7274(0.0818) | 1 |
| LLE | 0.9456(0.0251) | 0.7591(0.0454) | 0.7704(0.0757) | 0.9757(0.0076) | 0.5801(0.0771) | - | 0.7539(0.0426) | 0.7754(0.0797) | **0.9666(0.013)** | 0.7371(0.1082) | 1 |
| UMAP | 0.9369(0.0356) | 0.7135(0.0369) | 0.7704(0.0649) | 0.9728(0.0065) | 0.5566(0.0761) | - | 0.681(0.0435) | 0.6919(0.1052) | 0.9385(0.0333) | 0.6766(0.076) | 0 |
| Feature Selection | | | | | | | | | | | |
| CIFE | 0.9484(0.0229) | 0.7175(0.0564) | 0.737(0.075) | 0.9836(0.0059) | 0.6491(0.0616) | 0.4(0.2284) | 0.7123(0.0591) | 0.7862(0.0676) | 0.7188(0.1605) | 0.6984(0.1213) | 0 |
| CMIM | 0.937(0.0314) | 0.711(0.0624) | 0.7444(0.0508) | 0.9858(0.005) | 0.6516(0.0688) | **0.5667(0.2415)** | 0.7084(0.0582) | 0.749(0.0636) | 0.942(0.0222) | 0.6363(0.1152) | 1 |
| DISR | 0.9427(0.0297) | 0.7123(0.0585) | 0.7444(0.0591) | 0.9855(0.0052) | 0.6427(0.0776) | 0.5(0.3143) | 0.711(0.0615) | 0.7789(0.0621) | 0.9491(0.0254) | 0.6924(0.0743) | 0 |
| ICAP | 0.9384(0.0313) | 0.7148(0.0627) | 0.7519(0.0742) | 0.9858(0.005) | 0.6339(0.0881) | 0.4417(0.1669) | 0.711(0.0612) | 0.7603(0.0733) | 0.9473(0.0203) | 0.6016(0.0877) | 0 |
| JMI | 0.9413(0.0242) | 0.7084(0.0615) | 0.7593(0.0586) | 0.9861(0.005) | 0.6429(0.0662) | 0.375(0.1974) | 0.7123(0.058) | 0.7865(0.0582) | 0.9455(0.0211) | 0.6921(0.0856) | 0 |
| MIFS | 0.9556(0.0174) | **0.7773(0.043)** | 0.6519(0.1021) | 0.9842(0.0071) | 0.609(0.0578) | 0.3917(0.3094) | **0.7773(0.043)** | **0.8083(0.0921)** | 0.8998(0.0379) | 0.6618(0.061) | 3 |
| MIM | 0.9427(0.0275) | 0.7149(0.062) | 0.737(0.064) | 0.9858(0.005) | 0.6428(0.0698) | 0.45(0.1892) | 0.7109(0.0569) | 0.7637(0.0966) | 0.9403(0.0263) | 0.6766(0.0894) | 0 |
| MRMR | 0.9398(0.0312) | 0.776(0.0431) | 0.5778(0.0765) | **0.987(0.0064)** | 0.609(0.0578) | 0.3333(0.239) | **0.7773(0.0434)** | 0.7822(0.0937) | 0.5975(0.057) | 0.6618(0.061) | 2 |

# 10-fold Cross-Validation with Top 10 Features on 6 UCI Datasets

In addition to the 10-fold cross-validation experiments where the same features as TemporalPaD were selected in each fold, we further analyze the UCIH&M dataset from a different perspective by controlling the dimension of post-reduction. We select datasets from UCIH&M that had more than 10 features, totaling 6 datasets, and required all methods to reduce the original feature space to 10 dimensions (results are shown in Tables 8-13). Specifically, with the same number of features, we compare the performance of TemporalPaD against 18 feature dimensionality reduction algorithms, including 8 information-theory-based feature selection methods and 10 feature extraction algorithms. Following the comparison method described by [27], we evaluate the average Acc of three classifiers—SVM, NB, and C4.5—through 10-fold cross-validation as the number of selected features increased. Consistent with the approach taken by [27], we present the results by subtracting the average Acc of each baseline method from that of TemporalPaD. The column "#DiffRes>0 (Total 10)" indicates the count of times TemporalPaD outperform the current baseline method as the feature count increased from 1 to 10.

The analysis of Table 9 reveals that TemporalPaD's average Acc on the Spect dataset is notably superior to that of all 8 feature selection algorithms. This superiority stems from the fact that the Spect dataset exclusively comprises binary features [45], while TemporalPaD can nonlinearly map these features to a high-dimensional dense feature space through its representation module, a capability lacking in those feature selection algorithms. Similarly, the Lung-cancer dataset (refer to Table 13), which also contains comprehensive binary features, demonstrates similar characteristics. In the Heart dataset (refer to Table 12), which encompasses both categorical and continuous features, TemporalPaD still shows overall superiority over feature selection algorithms. However, despite outperforming individual feature extraction algorithms on these datasets,

TemporalPaD's overall performance is lower (see Tables 9, 12, and 13). This can be attributed to TemporalPaD's deep learning framework, which necessitates a larger sample size for model training compared to methods relying on statistical information [46], especially in cases like the Lung-cancer dataset where the number of features exceeds the sample size.

On the WPBC dataset (refer to Table 11), TemporalPaD exhibits exceptional performance compared to the majority of the 18 feature dimensionality reduction algorithms. This can be attributed to the presence of temporal features in the WPBC dataset [47], for which TemporalPaD is explicitly designed. In contrast, despite some overlap between the WDBC and WPBC datasets, TemporalPaD does not show notable improvement relative to feature extraction algorithms on the WDBC dataset, which lacks temporal features (refer to Table 10). Furthermore, our observations indicate that TemporalPaD's average Acc on the Hypothyroid dataset is unsatisfactory (refer to Table 8).

The above discussion highlights the significance of considering the distinct characteristics of the dataset and selecting an appropriate algorithm for feature dimensionality reduction [14]. Particularly, when dealing with datasets that exhibit temporal characteristics, TemporalPaD emerges as a more suitable choice compared to alternative approaches.

**Table 8:** The difference between the average Acc of TemporalPaD and those of 18 feature dimensionality reduction algorithms on the Hypothyroid dataset.

| Hypothyroid | 1 | 2 | 3 | 4 | 5 | 6 | 7 | 8 | 9 | 10 | #DiffRes>0(Total 10) |
|---|---|---|---|---|---|---|---|---|---|---|---|
| | | | | | Feature Selection | | | | | | |
| CIFE | -0.0924 | -0.0954 | -0.0952 | -0.1021 | -0.1062 | -0.1097 | -0.0933 | -0.1180 | -0.1416 | -0.1568 | 0 |
| CMIM | -0.0924 | -0.0954 | -0.0946 | -0.1014 | -0.1086 | -0.1151 | -0.0869 | -0.1092 | -0.1218 | -0.0922 | 0 |
| DISR | -0.0924 | -0.0831 | -0.0819 | -0.0765 | -0.0774 | -0.0752 | -0.0337 | -0.0543 | -0.0836 | -0.0209 | 0 |
| ICAP | -0.0924 | -0.0954 | -0.0946 | -0.1014 | -0.1086 | -0.1151 | -0.0869 | -0.1092 | -0.1218 | -0.0899 | 0 |
| JMI | -0.0924 | -0.0954 | -0.1116 | -0.1172 | -0.1175 | -0.1110 | -0.0869 | -0.1114 | -0.1398 | -0.0733 | 0 |
| MIFS | -0.0924 | -0.0954 | -0.0952 | -0.1070 | -0.0405 | -0.0280 | 0.0243 | 0.0135 | -0.0200 | -0.0400 | 2 |
| MIM | -0.0924 | -0.0954 | -0.1116 | -0.1170 | -0.1030 | -0.1151 | -0.0869 | -0.0598 | -0.0251 | -0.0400 | 0 |
| MRMR | -0.0924 | 0.1279 | -0.0928 | -0.1013 | -0.1023 | -0.1151 | -0.0924 | -0.1107 | -0.0401 | -0.0335 | 1 |
| | | | | | Feature Extraction | | | | | | |
| FA | 0.1862 | 0.1605 | -0.1006 | -0.1071 | -0.0691 | -0.0770 | -0.0669 | -0.0794 | -0.1305 | -0.1459 | 2 |
| PCA | -0.0886 | -0.0983 | -0.1137 | -0.1192 | -0.1302 | -0.1343 | -0.1188 | -0.1422 | -0.1734 | -0.1886 | 0 |
| KernelPCA | -0.0871 | -0.0995 | -0.1135 | -0.1193 | -0.1282 | -0.1318 | -0.1171 | -0.1414 | -0.1706 | -0.1854 | 0 |
| SparsePCA | -0.0891 | -0.0996 | -0.1084 | -0.1243 | -0.1307 | -0.1330 | -0.1159 | -0.1298 | -0.1591 | -0.1742 | 0 |
| SVD | -0.0880 | -0.0923 | -0.1116 | -0.1171 | -0.1277 | -0.1306 | -0.1140 | -0.1388 | -0.1687 | -0.1845 | 0 |
| ICA | -0.0886 | -0.0968 | -0.1152 | -0.1208 | -0.1282 | -0.1334 | -0.1180 | -0.1405 | -0.1711 | -0.1864 | 0 |
| NMF | -0.0880 | -0.0909 | -0.0949 | -0.1101 | -0.1287 | -0.1250 | -0.1066 | -0.1287 | -0.1571 | -0.1613 | 0 |
| ISOMAP | -0.0841 | -0.0878 | -0.1062 | -0.1172 | -0.1193 | -0.1215 | -0.1065 | -0.1308 | -0.1625 | -0.1783 | 0 |
| LLE | 0.0341 | 0.0442 | 0.0742 | 0.1136 | 0.0839 | 0.0782 | 0.0879 | 0.0886 | -0.1301 | -0.1458 | 8 |
| UMAP | -0.0860 | -0.0832 | -0.0960 | -0.1017 | -0.0983 | -0.1055 | -0.0860 | -0.1075 | -0.1308 | -0.1502 | 0 |

**Table 9:** The difference between the average Acc of TemporalPaD and those of 18 feature dimensionality reduction algorithms on the Spect dataset.

| Spect | 1 | 2 | 3 | 4 | 5 | 6 | 7 | 8 | 9 | 10 | #DiffRes>0(Total 10) |
|---|---|---|---|---|---|---|---|---|---|---|---|
| | | | | | Feature Selection | | | | | | |
| CIFE | 0.0000 | 0.0115 | 0.0500 | 0.0658 | 0.0247 | 0.0321 | 0.0275 | 0.0323 | 0.0298 | 0.0342 | 9 |
| CMIM | 0.0000 | 0.0115 | 0.0783 | 0.1347 | 0.0884 | 0.0847 | 0.0825 | 0.0800 | 0.0662 | 0.0452 | 9 |
| DISR | 0.0000 | 0.1162 | 0.0962 | 0.1170 | 0.1146 | 0.1158 | 0.0962 | 0.0921 | 0.0972 | 0.0689 | 9 |
| ICAP | 0.0000 | 0.0268 | 0.0748 | 0.1271 | 0.0833 | 0.0821 | 0.0774 | 0.0748 | 0.0599 | 0.0363 | 9 |
| JMI | 0.0000 | 0.0115 | 0.0686 | 0.0618 | 0.0385 | 0.0825 | 0.0639 | 0.0559 | 0.0659 | 0.0550 | 9 |
| MIFS | 0.0000 | 0.0244 | 0.1008 | 0.1081 | 0.0825 | 0.0865 | 0.0802 | 0.0751 | 0.0776 | 0.0501 | 9 |
| MIM | 0.0000 | 0.0128 | -0.0012 | 0.0023 | 0.0299 | 0.0745 | 0.0898 | 0.1010 | 0.1099 | 0.0850 | 8 |
| MRMR | 0.0000 | 0.0205 | 0.0896 | 0.1246 | 0.0835 | 0.0725 | 0.0692 | 0.0714 | 0.0776 | 0.0576 | 9 |
| | | | | | Feature Extraction | | | | | | |
| FA | -0.0191 | -0.0406 | -0.0387 | -0.0214 | -0.0223 | -0.0074 | -0.0311 | -0.0224 | -0.0226 | -0.0422 | 0 |
| PCA | -0.0151 | -0.0457 | -0.0450 | -0.0227 | -0.0349 | -0.0275 | -0.0348 | -0.0388 | -0.0349 | -0.0660 | 0 |
| KernelPCA | -0.0137 | -0.0572 | -0.0475 | -0.0277 | -0.0425 | -0.0364 | -0.0548 | -0.0413 | -0.0361 | -0.0523 | 0 |
| SparsePCA | -0.0151 | -0.0519 | -0.0287 | -0.0365 | -0.0314 | -0.0250 | -0.0373 | -0.0249 | -0.0138 | -0.0384 | 0 |
| SVD | -0.0099 | -0.0394 | -0.0347 | -0.0276 | -0.0350 | -0.0213 | -0.0347 | -0.0264 | -0.0338 | -0.0547 | 0 |
| ICA | -0.0151 | -0.0571 | -0.0374 | -0.0277 | -0.0364 | -0.0253 | -0.0298 | -0.0324 | -0.0189 | -0.0511 | 0 |
| NMF | -0.0099 | -0.0534 | -0.0288 | -0.0192 | -0.0225 | -0.0202 | -0.0399 | -0.0103 | -0.0151 | -0.0322 | 0 |
| ISOMAP | -0.0099 | -0.0534 | -0.0288 | -0.0192 | -0.0225 | -0.0202 | -0.0399 | -0.0103 | -0.0151 | -0.0322 | 0 |
| LLE | 0.0064 | -0.0370 | -0.0397 | -0.0252 | -0.0410 | -0.0253 | -0.0323 | -0.0263 | -0.0038 | -0.0260 | 1 |
| UMAP | 0.0328 | 0.0146 | 0.0089 | 0.0400 | 0.0240 | 0.0126 | 0.0194 | 0.0153 | 0.0330 | -0.0044 | 9 |

**Table 10:** The difference between the average Acc of TemporalPaD and those of 18 feature dimensionality reduction algorithms on the WDBC dataset.

| WDBC | 1 | 2 | 3 | 4 | 5 | 6 | 7 | 8 | 9 | 10 | #DiffRes>0(Total 10) |
|---|---|---|---|---|---|---|---|---|---|---|---|
| | **Feature Selection** | | | | | | | | | | |
| CIFE | 0.1451 | 0.0807 | 0.0612 | 0.0143 | 0.0336 | 0.0670 | 0.0739 | 0.0722 | 0.0639 | 0.0744 | 10 |
| CMIM | 0.1451 | 0.0842 | -0.0661 | -0.1118 | -0.0678 | -0.0275 | -0.0117 | -0.0163 | -0.0240 | -0.0099 | 2 |
| DISR | 0.1451 | -0.1381 | -0.1398 | -0.1334 | -0.0836 | -0.0514 | -0.0263 | -0.0268 | -0.0292 | -0.0187 | 1 |
| ICAP | 0.1451 | 0.0813 | -0.0638 | -0.1112 | -0.0666 | -0.0292 | -0.0082 | -0.0175 | -0.0240 | -0.0070 | 2 |
| JMI | 0.1451 | 0.0848 | 0.0246 | 0.0175 | -0.0460 | -0.0479 | -0.0310 | -0.0222 | -0.0321 | -0.0216 | 4 |
| MIFS | 0.1451 | -0.1381 | -0.1510 | -0.1504 | -0.1012 | -0.0562 | -0.0345 | -0.0398 | -0.0456 | -0.0345 | 1 |
| MIM | 0.1451 | 0.1661 | 0.1454 | 0.1442 | -0.0649 | -0.0274 | -0.0040 | -0.0263 | -0.0316 | -0.0163 | 4 |
| MRMR | 0.1451 | 0.1009 | -0.1510 | -0.1569 | -0.0977 | -0.0638 | -0.0374 | -0.0409 | -0.0415 | -0.0293 | 2 |
| | **Feature Extraction** | | | | | | | | | | |
| FA | -0.1888 | -0.1644 | -0.1521 | -0.1586 | -0.1046 | -0.0585 | -0.0374 | -0.0426 | -0.0468 | -0.0339 | 0 |
| PCA | -0.2094 | -0.1749 | -0.1732 | -0.1744 | -0.1158 | -0.0690 | -0.0462 | -0.0444 | -0.0497 | -0.0351 | 0 |
| KernelPCA | -0.2123 | -0.1820 | -0.1680 | -0.1639 | -0.1105 | -0.0708 | -0.0456 | -0.0468 | -0.0509 | -0.0374 | 0 |
| SparsePCA | -0.2106 | -0.1772 | -0.1679 | -0.1698 | -0.1135 | -0.0632 | -0.0444 | -0.0509 | -0.0421 | -0.0328 | 0 |
| SVD | -0.2030 | -0.1767 | -0.1703 | -0.1727 | -0.1076 | -0.0661 | -0.0445 | -0.0444 | -0.0456 | -0.0369 | 0 |
| ICA | -0.2094 | -0.1691 | -0.1650 | -0.1598 | -0.1023 | -0.0614 | -0.0280 | -0.0345 | -0.0251 | -0.0029 | 0 |
| NMF | -0.2030 | -0.1685 | -0.1627 | -0.1721 | -0.1146 | -0.0702 | -0.0409 | -0.0480 | -0.0439 | -0.0293 | 0 |
| ISOMAP | -0.2135 | -0.1802 | -0.1773 | -0.1756 | -0.1129 | -0.0655 | -0.0433 | -0.0403 | -0.0450 | -0.0292 | 0 |
| LLE | -0.2164 | -0.1855 | -0.1756 | -0.1692 | -0.1012 | -0.0562 | -0.0345 | -0.0339 | -0.0345 | -0.0252 | 0 |
| UMAP | -0.2036 | -0.1761 | -0.1709 | -0.1610 | -0.1035 | -0.0585 | -0.0228 | -0.0362 | -0.0316 | -0.0217 | 0 |

**Table 11:** The difference between the average Acc of TemporalPaD and those of 18 feature dimensionality reduction algorithms on the WPBC dataset.

| WPBC | 1 | 2 | 3 | 4 | 5 | 6 | 7 | 8 | 9 | 10 | #DiffRes>0(Total 10) |
|---|---|---|---|---|---|---|---|---|---|---|---|
| | **Feature Selection** | | | | | | | | | | |
| CIFE | 0.0056 | -0.0369 | -0.0193 | -0.0236 | -0.0016 | -0.0101 | -0.0105 | -0.0297 | -0.0282 | -0.0112 | 1 |
| CMIM | 0.0056 | -0.0403 | -0.0016 | -0.0225 | -0.0005 | 0.0068 | -0.0199 | -0.0110 | -0.0111 | 0.0275 | 3 |
| DISR | 0.0056 | -0.0133 | -0.0055 | -0.0259 | 0.0166 | -0.0055 | -0.0089 | -0.0149 | -0.0188 | 0.0132 | 3 |
| ICAP | 0.0056 | -0.0419 | 0.0036 | -0.0308 | 0.0046 | 0.0018 | -0.0132 | -0.0194 | -0.0062 | 0.0310 | 5 |
| JMI | 0.0056 | -0.0385 | 0.0105 | -0.0122 | 0.0318 | 0.0350 | 0.0177 | 0.0087 | 0.0219 | 0.0471 | 8 |
| MIFS | 0.0056 | -0.0133 | 0.0017 | -0.0254 | 0.0081 | -0.0086 | -0.0369 | -0.0480 | -0.0430 | -0.0297 | 3 |
| MIM | 0.0056 | -0.0101 | 0.0097 | -0.0155 | 0.0167 | 0.0198 | 0.0045 | 0.0004 | 0.0154 | 0.0304 | 8 |
| MRMR | 0.0056 | -0.0133 | 0.0016 | -0.0254 | 0.0068 | -0.0215 | -0.0452 | -0.0311 | -0.0463 | -0.0246 | 3 |
| | **Feature Extraction** | | | | | | | | | | |
| FA | -0.0163 | -0.0318 | 0.0136 | -0.0207 | 0.0117 | 0.0147 | -0.0024 | 0.0034 | -0.0067 | 0.0443 | 5 |
| PCA | -0.0029 | -0.0387 | -0.0139 | -0.0478 | -0.0155 | -0.0238 | -0.0286 | -0.0348 | -0.0447 | -0.0332 | 0 |
| KernelPCA | 0.0008 | -0.0271 | -0.0023 | -0.0289 | -0.0002 | -0.0254 | -0.0183 | -0.0263 | -0.0430 | -0.0196 | 1 |
| SparsePCA | -0.0211 | -0.0368 | 0.0167 | -0.0607 | -0.0204 | -0.0169 | -0.0287 | -0.0329 | -0.0231 | -0.0063 | 1 |
| SVD | -0.0045 | -0.0405 | -0.0022 | -0.0326 | -0.0004 | -0.0339 | -0.0303 | -0.0281 | -0.0483 | -0.0300 | 0 |
| ICA | -0.0029 | -0.0099 | 0.0028 | -0.0356 | -0.0072 | -0.0223 | -0.0271 | -0.0265 | -0.0508 | -0.0250 | 1 |
| NMF | -0.0045 | -0.0150 | 0.0336 | -0.0325 | 0.0150 | -0.0454 | -0.0273 | -0.0332 | -0.0350 | -0.0315 | 2 |
| ISOMAP | -0.0129 | -0.0332 | 0.0304 | -0.0038 | 0.0454 | 0.0468 | 0.0268 | 0.0324 | 0.0343 | 0.0423 | 7 |
| LLE | -0.0417 | -0.0033 | 0.0303 | 0.0116 | 0.0520 | 0.0331 | 0.0181 | 0.0241 | 0.0173 | 0.0304 | 8 |
| UMAP | -0.0148 | 0.0106 | 0.0505 | 0.0211 | 0.0405 | 0.0404 | 0.0364 | 0.0291 | 0.0394 | 0.0534 | 9 |

**Table 12:** The difference between the average Acc of TemporalPaD and those of 18 feature dimensionality reduction algorithms on the Heart dataset.

| Heart | 1 | 2 | 3 | 4 | 5 | 6 | 7 | 8 | 9 | 10 | #DiffRes>0(Total 10) |
|---|---|---|---|---|---|---|---|---|---|---|---|
| | Feature Selection | | | | | | | | | | |
| CIFE | 0.0914 | 0.1074 | 0.0000 | 0.0506 | 0.0136 | 0.0062 | 0.0099 | 0.0173 | -0.0111 | -0.0111 | 7 |
| CMIM | 0.0914 | 0.1086 | 0.0111 | 0.0407 | -0.0963 | -0.0716 | -0.0407 | -0.0259 | -0.0099 | -0.0123 | 4 |
| DISR | 0.0914 | 0.1086 | 0.0370 | 0.0420 | -0.0963 | -0.0765 | -0.0519 | -0.0407 | -0.0099 | -0.0198 | 4 |
| ICAP | 0.0914 | -0.0111 | 0.0185 | 0.0383 | -0.0926 | -0.0704 | -0.0370 | -0.0346 | -0.0123 | -0.0136 | 3 |
| JMI | 0.0914 | 0.1074 | 0.0074 | 0.0284 | -0.0963 | -0.0543 | -0.0358 | -0.0259 | -0.0062 | -0.0123 | 4 |
| MIFS | 0.0914 | -0.0185 | 0.0222 | 0.0062 | -0.0296 | -0.0235 | -0.0259 | 0.0037 | 0.0123 | 0.0086 | 6 |
| MIM | 0.0914 | -0.0123 | 0.0000 | 0.0012 | -0.0938 | -0.0753 | -0.0568 | -0.0333 | -0.0111 | -0.0086 | 2 |
| MRMR | 0.0914 | 0.0642 | 0.0049 | 0.0309 | -0.0593 | -0.0778 | -0.0420 | -0.0272 | 0.0012 | -0.0333 | 5 |
| | Feature Extraction | | | | | | | | | | |
| FA | -0.1346 | -0.1259 | -0.1062 | -0.0568 | -0.0901 | -0.0827 | -0.0630 | -0.0407 | -0.0049 | -0.0259 | 0 |
| PCA | -0.1568 | -0.1457 | -0.1123 | -0.0728 | -0.0840 | -0.0790 | -0.0741 | -0.0593 | -0.0272 | -0.0272 | 0 |
| KernelPCA | -0.1481 | -0.1420 | -0.1086 | -0.0630 | -0.0790 | -0.0815 | -0.0667 | -0.0593 | -0.0259 | -0.0333 | 0 |
| SparsePCA | -0.1481 | -0.1321 | -0.0914 | -0.0741 | -0.0691 | -0.0765 | -0.0543 | -0.0395 | -0.0309 | -0.0222 | 0 |
| SVD | -0.1383 | -0.1346 | -0.1160 | -0.0679 | -0.0926 | -0.0852 | -0.0679 | -0.0556 | -0.0370 | -0.0309 | 0 |
| ICA | -0.1568 | -0.1321 | -0.1049 | -0.0605 | -0.0691 | -0.0630 | -0.0593 | -0.0321 | -0.0074 | -0.0049 | 0 |
| NMF | -0.1383 | -0.1173 | -0.0963 | -0.0654 | -0.0691 | -0.0753 | -0.0519 | -0.0210 | -0.0284 | -0.0148 | 0 |
| ISOMAP | -0.1630 | -0.1222 | -0.1000 | -0.0617 | -0.0827 | -0.0840 | -0.0667 | -0.0370 | -0.0185 | -0.0198 | 0 |
| LLE | -0.0531 | -0.1012 | -0.0679 | -0.0469 | -0.0630 | -0.0654 | -0.0605 | -0.0296 | -0.0222 | -0.0185 | 0 |
| UMAP | -0.1432 | -0.1296 | -0.1111 | -0.0556 | -0.0667 | -0.0864 | -0.0543 | -0.0395 | -0.0185 | -0.0062 | 0 |

**Table 13:** The difference between the average Acc of TemporalPaD and those of 18 feature dimensionality reduction algorithms on the Lung-cancer dataset.

| Lung-cancer | 1 | 2 | 3 | 4 | 5 | 6 | 7 | 8 | 9 | 10 | #DiffRes>0(Total 10) |
|---|---|---|---|---|---|---|---|---|---|---|---|
| | Feature Selection | | | | | | | | | | |
| CIFE | 0.0028 | 0.0861 | -0.0056 | 0.0222 | 0.0111 | 0.0472 | -0.0028 | -0.0278 | 0.0667 | 0.0417 | 7 |
| CMIM | 0.0028 | 0.0861 | -0.1222 | -0.0306 | 0.0167 | 0.0222 | -0.0167 | -0.0806 | -0.1056 | -0.1222 | 4 |
| DISR | 0.0028 | -0.1194 | -0.1639 | -0.1917 | -0.1556 | -0.1194 | -0.1833 | -0.2167 | -0.1222 | -0.1917 | 1 |
| ICAP | 0.0028 | 0.0444 | -0.0889 | -0.1028 | -0.0694 | 0.0278 | -0.0139 | -0.0722 | 0.0250 | -0.0361 | 4 |
| JMI | 0.0028 | 0.0861 | -0.0194 | -0.0194 | -0.0194 | 0.0167 | -0.0639 | -0.0222 | -0.0500 | -0.1194 | 3 |
| MIFS | 0.0028 | -0.1083 | -0.0389 | -0.0667 | -0.0528 | 0.0528 | -0.0222 | -0.0556 | 0.0056 | -0.0500 | 3 |
| MIM | 0.0028 | -0.1083 | -0.2000 | -0.1833 | -0.1194 | -0.1000 | -0.1278 | -0.1778 | -0.1639 | -0.1917 | 1 |
| MRMR | 0.0028 | -0.0917 | -0.0417 | -0.0639 | -0.0667 | -0.0472 | -0.1000 | -0.1528 | -0.1889 | -0.2083 | 1 |
| | Feature Extraction | | | | | | | | | | |
| FA | -0.1056 | 0.0139 | -0.1500 | -0.0833 | 0.0472 | 0.1028 | 0.0500 | 0.1000 | -0.0222 | -0.0556 | 5 |
| PCA | -0.1889 | -0.1167 | -0.2167 | -0.2111 | -0.1583 | -0.0889 | -0.1583 | -0.1750 | -0.1556 | -0.1611 | 0 |
| KernelPCA | -0.2083 | -0.1167 | -0.2250 | -0.1806 | -0.1500 | -0.0806 | -0.1583 | -0.1611 | -0.0833 | -0.1250 | 0 |
| SparsePCA | -0.1917 | -0.0972 | -0.2111 | -0.1556 | -0.1139 | -0.0361 | -0.0833 | -0.1639 | -0.0972 | -0.1056 | 0 |
| SVD | -0.0167 | -0.1000 | -0.1444 | -0.1333 | -0.1028 | -0.0389 | -0.0778 | -0.1028 | -0.0889 | -0.1111 | 0 |
| ICA | -0.1889 | -0.0528 | -0.2000 | -0.1639 | -0.0806 | -0.0333 | -0.0556 | -0.0583 | -0.0083 | -0.0694 | 0 |
| NMF | -0.0167 | -0.0722 | -0.1917 | -0.1639 | -0.0667 | -0.0278 | -0.0222 | -0.1194 | 0.0361 | -0.1000 | 1 |
| ISOMAP | -0.1583 | -0.0750 | -0.0972 | -0.1056 | -0.0556 | 0.0361 | 0.0000 | -0.1000 | -0.0583 | -0.0389 | 1 |
| LLE | -0.0944 | -0.0056 | -0.0750 | -0.0806 | -0.0111 | 0.0111 | -0.0722 | -0.0417 | 0.0361 | -0.0389 | 2 |
| UMAP | 0.1306 | -0.0528 | -0.1472 | -0.1028 | -0.0250 | 0.0028 | 0.0083 | -0.0944 | -0.0528 | -0.0500 | 3 |

## Ablation Study Based on the Breast Dataset of UCI

We conduct an ablation study to evaluate the effectiveness of feature reduction learned through the policy module, i.e., reinforcement learning, using the Breast dataset [48] as a case study.

To perform the evaluation, the Breast dataset is randomly divided into training and test sets with a 9:1 ratio. A fixed seed of 75 is used to ensure the reproducibility and consistency as the above experiments. The results are presented in Figure 2, where the horizontal axis represents the various training phases of TemporalPaD: '1' denotes the pre-training phase of the representation module, '2' represents the pre-training phase of the policy module, and '3' indicates the joint training phase.

From the results, it can be observed that after integrating the policy module with the representation module, the Acc of TemporalPaD on the training dataset increased by 0.0017, from 0.9683 to 0.97, while maintaining Acc on the test set. This improvement in Acc on the Breast dataset demonstrates TemporalPaD's capability to extract and reduce features effectively from the entire feature set. This result further supports the concept that incorporating reinforcement learning into TemporalPaD through the policy module has the potential to enhance the performance of downstream tasks.

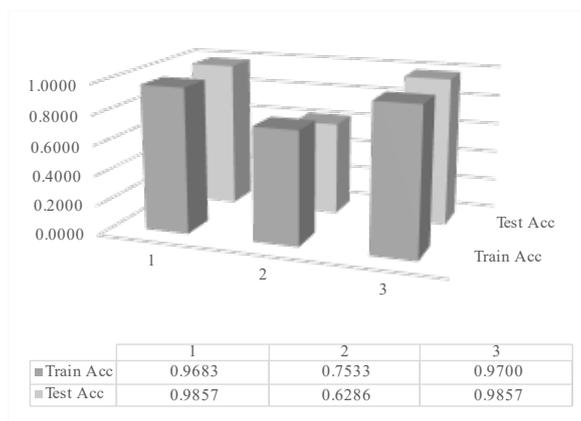

Figure 2. The Acc of TemporalPaD in three training phases on the test set of the Breast dataset.

## A Real Application of TemporalPaD on DNA Sequences

TemporalPaD has been extensively evaluated alongside various feature reduction algorithms using the UCI dataset, a well-known feature reduction benchmark. The results indicate that TemporalPaD is particularly well-suited for time series data, attributed to its representation module designed with embeddings and LSTM. Building on this, we apply TemporalPaD to a real-world problem: the binary classification of DNA sequences.

As mentioned earlier, there is a public dataset for the enhancer classification task, referenced as Enhancer1 and Enhancer2 in Table 1. These correspond to two classification tasks: the enhancer category task and the enhancer strength task. Both tasks are binary classifications, where Enhancer1 includes enhancers (labeled 1) and non-enhancers (labeled 0), while Enhancer2 contains strong enhancers (labeled 1) and weak enhancers (labeled 0).

To successfully implement TemporalPaD, we treat the DNA sequences as natural language. This approach allows us to encode DNA sequences and extract features for contextual dependencies. Subsequently, TemporalPaD is employed to reduce these features. During this process, we use Enhancer1, the enhancer category classification task, as a reference to select appropriate hyperparameters for TemporalPaD. Specifically, we split the training dataset of Enhancer1 into a training subset and a validation subset in a 9:1 ratio to determine the best hyperparameters, which yield the highest accuracy (Acc) on the validation dataset during the training phase of TemporalPaD.

Based on the optimal hyperparameters determined from Enhancer1, we then train TemporalPaD on both Enhancer1 and Enhancer2 datasets using the same hyperparameters. Finally, we compare the performance of TemporalPaD with several classification enhancer classifiers.

## Kmer

In this section, we utilize the kmer technology for segmentation, which allows us to transform DNA sequences into a suitable format for feature extraction. The kmer technique partitions the sequences into smaller fragments of length k, as depicted in

Figure 3. For an original sequence of length n-bp, we employ k-length tuples (kmers) sequentially to divide the sequence, resulting in n-k+1 kmer fragments. By processing all samples of the training dataset in this manner, integrating these fragments, and deduplicating them, we form a mapping table. This table is sorted according to the order of appearance to generate a mapping from the string segments to real numbers.

Using the mapping relationship between the kmers and real numbers, we obtain the encoded sequence, EncodeSeq. This enables us to derive the corresponding encoded feature representation through embedding techniques. This process converts the positional information into meaningful dense representations that can be further analyzed using the feature extraction layer.

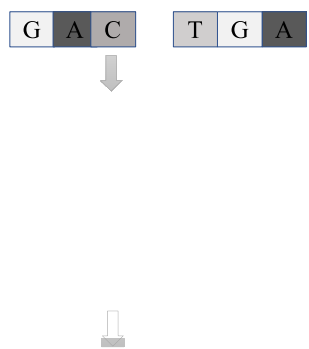

**Figure 3.** Kmer sequence segmentation illustration.

## Selection of Reward Calculation Method

The selection of an appropriate reward calculation method is crucial for the performance of RL, as specified in formula (9) of TemporalPaD. To determine the most suitable reward computation method for the enhancer classification task, we conduct an experiment comparing the performance of three different reward calculations from formula (9), while keeping other configurations of TemporalPaD constant, based on the first enhancer task: enhancer category classification.

In this experiment, the kmer parameter is fixed at 1, treating each position in the sequence as an individual feature. The policy module is designed as a single-layer fully connected network, while the representation module utilizes a single-layer unidirectional LSTM network. Acc is chosen as the evaluation metric to assess the performance of the three different reward calculations.

By comparing the Acc of the three reward computations on the enhancer category task (depicted in Figure 4), we find that the second reward computation method, i.e., A(2) from formula (9), shows the highest Acc on the validation set of Enhancer1. Consequently, we select the second reward calculation method for all subsequent experiments.

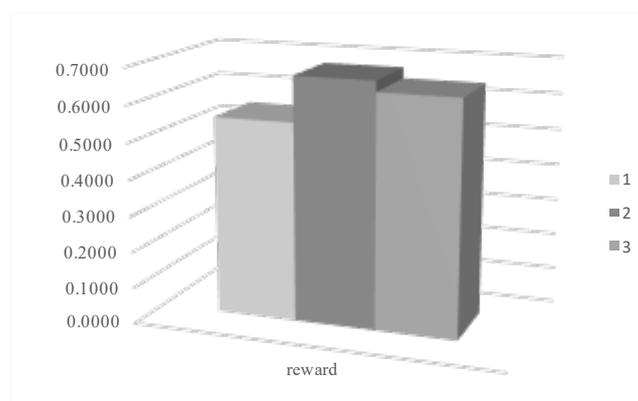

**Figure 4.** Fixing the k value of Kmer, comparison of the Acc for three different reward calculations on the validation dataset of Enhancer1. '1' refers to A(1) of formula (9); '2' denotes A(2) of formula (9); and '3' represents A(3) of formula (9).

# Evaluation of Network Architecture for Policy Module and Representation Module

TemporalPaD is a theoretical framework that integrates RL with neural networks to conduct feature reduction from a temporal perspective. In this section, we compare different neural network configurations for the Policy Module and Representation Module in TemporalPaD, setting the kmer parameter to 1 and using A(2) from formula (9) for reward calculation.

For the Policy Module, we compare three options: a one-layer fully connected network, a two-layer fully connected network, and textCNN [19]. These alternative network configurations allow us to evaluate TemporalPaD's performance under different Policy Module structures.

Additionally, for the Representation Module, which plays a pivotal role in capturing temporal dependencies in time series data, we analyze three primary network structures: a one-layer unidirectional LSTM, a one-layer bidirectional LSTM (Bi-LSTM), and a two-layer Bi-LSTM. By assessing TemporalPaD's Acc on the validation dataset of Enhancer1 with these various Representation Module configurations, we gain insights into their effectiveness for enhancer category classification.

Next, we traverse all combinations of the Policy Module with the Representation Module and evaluate their Acc on the validation dataset of Enhancer1 to determine the most suitable combinations. The results are shown in Figure 5.

(a) Fixed Representation Module as one-layer LSTM: Policy Module as one-layer fully connected network (a1), two-layer fully connected network (a2), and textCNN (a3).
(b) Fixed Representation Module as one-layer Bi-LSTM: Policy Module as one-layer fully connected network (b1), two-layer fully connected network (b2), and textCNN (b3).
(c) Fixed Representation Module as two-layer Bi-LSTM: Policy Module as one-layer fully connected network (c1), two-layer fully connected network (c2), and textCNN (c3).

The results, as shown in Figure 5, indicate that the network configuration with a one-layer LSTM as the Representation Module and a two-layer fully connected network as the Policy Module achieves the highest Acc on the validation dataset of Enhancer1 when k of kmer is fixed at 1 and A(2) of formula (9) is used for reward computation. Additionally, comparing configurations (a1), (a2), and (c1) supports previous research indicating that the Policy Module's network should not be excessively complex [17].

Therefore, we select combination 'a2' as the optimal configuration for the Policy Module and Representation Module of TemporalPaD for subsequent experiments. This configuration strikes a balance between model complexity and performance.

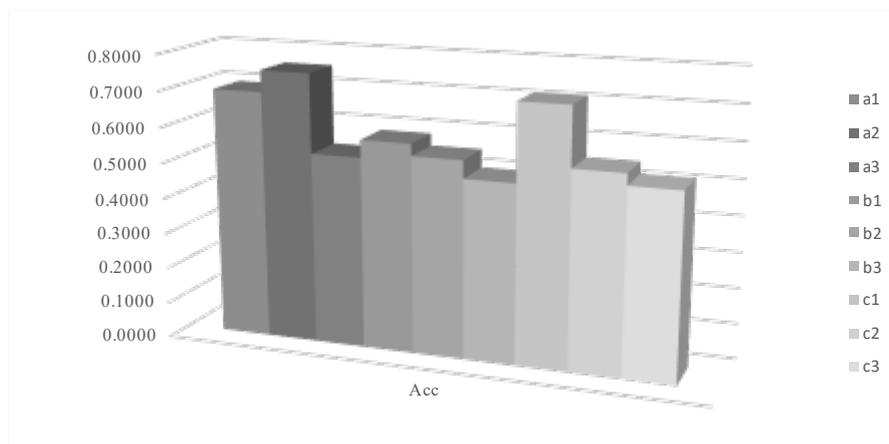

**Figure 5.** Acc of different combinations of Policy Module and Representation Module in TemporalPaD on the validation dataset of Enhancer1. 'a1', 'a2', and 'a3' represent a one-layer fully connected network, a two-layer fully connected network, and textCNN for the Policy Module, respectively, with the Representation Module as a one-layer unidirectional LSTM. 'b1', 'b2', and 'b3' represent a one-layer fully connected network, a two-layer fully connected network, and textCNN for the Policy Module, respectively, with the Representation Module as a one-layer Bi-LSTM. 'c1', 'c2', and 'c3' represent a one-layer fully connected network, a two-layer fully connected network, and textCNN for the Policy Module, respectively, with the Representation Module as a two-layer Bi-LSTM.

# Selection of K for Kmer

In the previous section, we determined the appropriate reward calculation and specific network structures of TemporalPaD for solving the enhancer category classification. In this section, our focus shifts to selecting the optimal k value for kmer in both enhancer tasks: enhancer category (Enhancer1) and enhancer strength classification (Enhancer2), with accuracy (Acc) as our primary consideration. We explore k values for kmer ranging from 1 to 6. The results are shown in Figure 6. The '1 layer' of Figure 6 illustrates the Acc of TemporalPaD for different kmer values in the enhancer category classification, while the '2 layer' displays the Acc of TemporalPaD for different kmer values in the enhancer strength classification.

The results in Figure 6 consistently demonstrate the superiority of kmer = 1 in both tasks. For enhancer category classification, kmer = 1 consistently outperforms other settings, achieving an accuracy of 73% in training stage 3 of TemporalPaD, representing an 8% improvement over the second-best setting, kmer = 2. Similarly, in the enhancer strength task, kmer = 1 achieves an accuracy of 62.4% in training stage 3 of TemporalPaD, surpassing kmer = 2 by 6.93%. These results, obtained from the validation set, provide robust evidence that kmer = 1 is the optimal parameter for conducting enhancer classification with TemporalPaD and for comparing the results with existing classical methods.

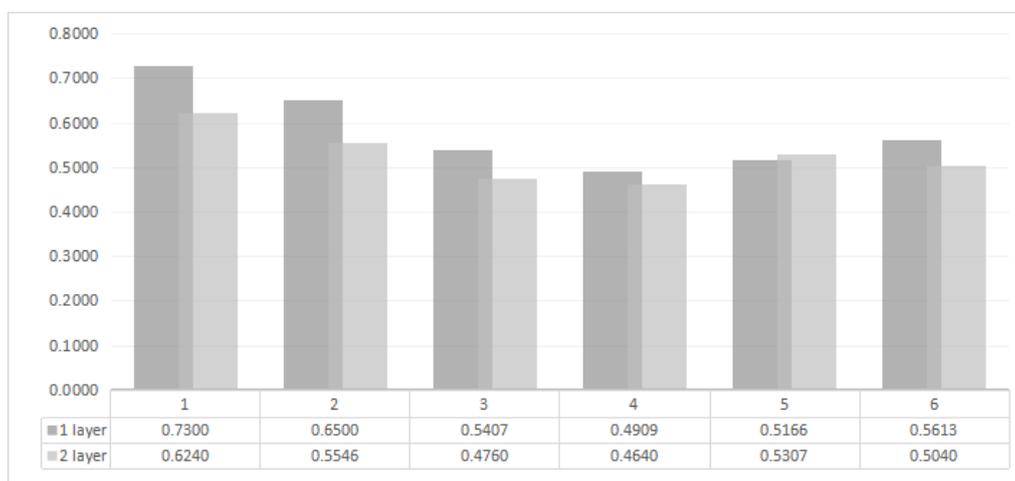

**Figure 6.** Acc of different kmer values on the validation set. '1 layer' denotes the enhancer category classification (Enhancer1), while '2 layer' denotes the enhancer strength classification (Enhancer2).

## Comparison with Classical Methods

In this section, we compare the performance of TemporalPaD against three existing classical enhancer classifiers using the same independent test set after training on the same training dataset. The results of this comparison are presented in Table 14, showcasing the performance of TemporalPaD across five different metrics.

Given that sensitivity (Sn) and specificity (Sp) metrics offer different perspectives in binary models, improvements in one metric may correspond to declines in the other [49, 50]. Guided by this principle, our primary focus is to evaluate the performance of TemporalPaD in both enhancer classification tasks using Acc and Matthews correlation coefficient (MCC) [24] to reflect overall performance.

In the enhancer category classification, TemporalPaD achieves an Acc of 0.7575, surpassing the second-best method by 1%, and exceeds the second-best by 2.79% in terms of MCC. For strength classification, TemporalPaD outperforms the second-best method by 2.5% in terms of Acc. Notably, in the strength classification task, TemporalPaD surpasses the second-best classifier by 18.03% in the context of MCC. These results provide robust evidence of the effectiveness and competitive performance of TemporalPaD in the dimensionality reduction of temporal data.

**Table 14.** Comparison of TemporalPaD with the three classical classification models on the independent dataset.

|  | Methods | Acc | MCC | Sn | Sp | AUC |
|---|---|---|---|---|---|---|
| enhancers vs non-enhancers | TemporalPaD (Ours) | 0.7575 | 0.5243 | 0.7750 | 0.7400 | 0.7575 |
|  | iEnhancer-EL | 0.7475 | 0.4964 | 0.7100 | 0.7850 | 0.8173 |
|  | iEnhancer-2L | 0.7300 | 0.4604 | 0.7100 | 0.7500 | 0.8062 |
|  | EnhancerPred | 0.7400 | 0.4800 | 0.7350 | 0.7450 | 0.8013 |
| strong enhancers vs weak enhancers | TemporalPaD (Ours) | 0.6350 | 0.4025 | 0.9100 | 0.3600 | 0.6350 |
|  | iEnhancer-EL | 0.6100 | 0.2222 | 0.5400 | 0.6800 | 0.6801 |
|  | iEnhancer-2L | 0.6050 | 0.2181 | 0.4700 | 0.7400 | 0.6678 |
|  | EnhancerPred | 0.5500 | 0.1021 | 0.4500 | 0.6500 | 0.5790 |

## Ablation Study of TemporalPaD Conducted on Super-Enhancer Prediction

To further validate the efficiency of TemporalPaD, a deep learning framework integrating RL with neural networks, we expand the original independent dataset of Enhancer2 (enhancer strength classification) to demonstrate the necessity of the policy module within TemporalPaD. We focus on super-enhancers (SEs), a class of strong enhancers known for their association with potent enhancers [51].

SEs are cell-specific cis-regulatory elements in DNA that play a critical role in the precise regulation of downstream genes. We selected 78 SEs from [51] database, which exhibited a similarity score greater than 75% to the strong enhancers in the independent dataset of Enhancer2. Subsequently, we utilize the three-phase TemporalPaD, trained on the training dataset of Enhancer2, to predict the super-enhancers for these 78 SEs, as illustrated in Figure 7.

As shown in Figure 7, the initial Acc at training stage 1 of TemporalPaD is modest, at 0.4133. However, after integrating the policy module, which uses RL for feature reduction, we observe a significant improvement in Acc, reaching 0.6400 at training stage 3 (marked as '3' in Figure 7). This notable enhancement validates the practical applicability of our policy module in effectively categorizing the strength of enhancers. This finding further substantiates TemporalPaD as a proficient framework for feature dimension reduction in time series data. When coupled with appropriate classification models, it enhances the performance of subsequent models.

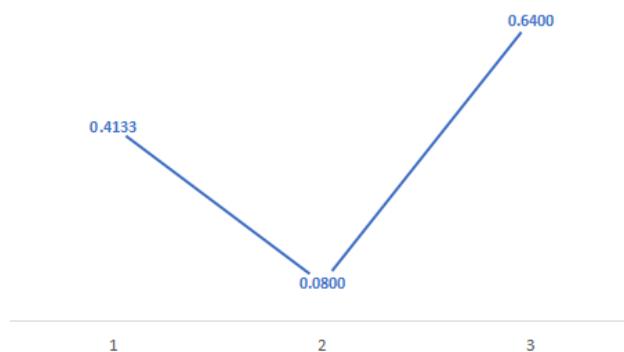

**Figure 7.** The Acc of TemporalPaD on the SEs dataset throughout the three training stages of TemporalPaD. The x-axis corresponds to the three training stages of TemporalPaD, designated as '1', '2', and '3'. The y-axis represents the Acc.

# Conclusion

In this study, we propose a novel end-to-end deep learning framework called TemporalPaD, which integrates RL with neural networks. The framework consists of a Representation Module for feature extraction and a Policy Module for dimensionality reduction through RL. We comprehensively evaluate TemporalPaD from various perspectives using the UCI dataset, a well-known benchmark for validating feature reduction algorithms, including 10 independent tests and 10-fold cross-validation.

Beyond these evaluations, considering that TemporalPaD is specifically designed for time series data, we applied it to a real-world DNA classification problem: enhancer category and enhancer strength. The results demonstrate that TemporalPaD is an efficient and effective framework for achieving feature reduction, whether on structured data or sequence datasets.

However, we also recognize opportunities for further improvement of TemporalPaD. For example, introducing multi-agent systems could result in faster convergence and enhanced performance.

## Funding

This work was supported by the Senior and Junior Technological Innovation Team (20210509055RQ), Guizhou Provincial Science and Technology Projects (ZK2023-297), the Science and Technology Foundation of Health Commission of Guizhou Province (gzwkj2023-565), Science and Technology Project of Education Department of Jilin Province (JJKH20220245KJ and JJKH20220226SK), the National Natural Science Foundation of China (62072212 and U19A2061), the Jilin Provincial Key Laboratory of Big Data Intelligent Computing (20180622002JC), and the Fundamental Research Funds for the Central Universities, JLU.